%% file: main.tex
\documentclass[10pt,twocolumn,letterpaper]{article}

\usepackage[pagenumbers]{cvpr}

\input{preamble}

\definecolor{cvprblue}{rgb}{0.21,0.49,0.74}
\usepackage[pagebackref,breaklinks,colorlinks,allcolors=cvprblue]{hyperref}

\newcommand{\authorskip}{\hspace{8mm}}
\newcommand{\institutionskip}{\hspace{10mm}}

\title{In Pursuit of Pixel Supervision for Visual Pre-training}

\author{
Lihe Yang\textsuperscript{2,1\thanks{This work was done during an internship at Meta.}}\authorskip
Shang-Wen Li\textsuperscript{1}\authorskip
Yang Li\textsuperscript{1}\authorskip
Xinjie Lei\textsuperscript{1}\authorskip\\
Dong Wang\textsuperscript{1}\authorskip
Abdelrahman Mohamed\textsuperscript{1}\authorskip
Hengshuang Zhao\textsuperscript{2}\authorskip
Hu Xu\textsuperscript{1}\vspace{1mm}\\
\textsuperscript{1}FAIR, Meta\institutionskip
\textsuperscript{2}HKU\\
{\tt \small \url{https://github.com/facebookresearch/pixio}}
}

\begin{document}

\maketitle

\begin{abstract}
\input{sec/0_abstract}
\end{abstract}
\input{sec/1_intro}
\input{sec/2_related_work}
\input{sec/3_method}
\input{sec/4_exp}
\input{sec/6_conclusion}

\clearpage
\clearpage
\newpage

\section*{Acknowledgments}
We thank Kaiming He, Saining Xie, Lifei Huang, Ramya Raghavendra, Stephane Kasriel, and Rob Fergus for their helpful discussions and valuable feedback.

{
    \small
    \bibliographystyle{ieeenat_fullname}
    \bibliography{main}
}

\input{sec/X_suppl}

\end{document}

%% file: preamble.tex
\usepackage[dvipsnames]{xcolor}
\usepackage{multirow}
\usepackage{xspace}
\usepackage{enumitem}

\makeatletter
\DeclareRobustCommand\onedot{\futurelet\@let@token\@onedot}
\def\@onedot{\ifx\@let@token.\else.\null\fi\xspace}

\def\eg{\emph{e.g}\onedot} 
\def\ie{\emph{i.e}\onedot} 
 
\def\etc{\emph{etc}\onedot}

\makeatother

\newcommand{\method}{Pixio\xspace}

\usepackage{pifont}
\newcommand{\cmark}{\ding{51}}
\newcommand{\xmark}{\ding{55}}

%% file: sec/0_abstract.tex
At the most basic level, pixels are the source of the visual information through which we perceive the world.
Pixels contain information at all levels, ranging from low-level attributes to high-level concepts. Autoencoders represent a classical and long-standing paradigm for learning representations from pixels or other raw inputs. In this work, we demonstrate that autoencoder-based self-supervised learning remains competitive today and can produce strong representations for downstream tasks, while remaining simple, stable, and efficient. Our model, codenamed ``\method'', is an enhanced masked autoencoder (MAE) with more challenging pre-training tasks and more capable architectures.
The model is trained on 2B web-crawled images with a self-curation strategy with minimal human curation. \method performs competitively across a wide range of downstream tasks in the wild, including monocular depth estimation (\eg, Depth Anything), feed-forward 3D reconstruction (\ie, MapAnything), semantic segmentation, and robot learning, outperforming or matching DINOv3 trained at similar scales. Our results suggest that pixel-space self-supervised learning can serve as a promising alternative and a complement to latent-space approaches.

%% file: sec/1_intro.tex
\section{Introduction}

Over the past decade, progress in computer vision has consistently been driven by corresponding advances in representation learning.
Starting from supervised representation learning~\cite{alexnet} based on human annotations (\eg, ImageNet~\cite{imagenet}), the vision community has since transitioned to self-supervised representation learning (\eg, \cite{moco, simclr, mae, dino}) using unannotated data.
The rich information contained in the data itself can serve as a powerful source of supervision for learning general-purpose representations.

Modern methods on self-supervised learning generally fall into two categories, depending on where the objective is formulated: the \textit{raw} input space (\ie, pixels in the case of vision data) or a \textit{latent} space produced by models. The first category is represented by denoising autoencoders (DAE)~\cite{vincent2008extracting}, now commonly realized as masked autoencoders (MAE)~\cite{mae} which learn by predicting unknown pixels under structural corruptions. 
The second category stems from contrastive learning~\cite{Hadsell2006, moco, simclr} and has evolved to incorporate various forms of latent-space objectives (\eg, DINO~\cite{dino}, JEPA~\cite{jepa}). Today, the go-to solution for off-the-shelf self-supervised learning models is typically DINO and its extensions, whereas MAEs often serve as pre-trained initializations for fine-tuning foundation models~\cite{sam1,sam2,d4rt}.

\begin{figure*}[t]
    \centering
    \begin{subfigure}{0.494\linewidth}
        \centering
        \includegraphics[width=\linewidth]{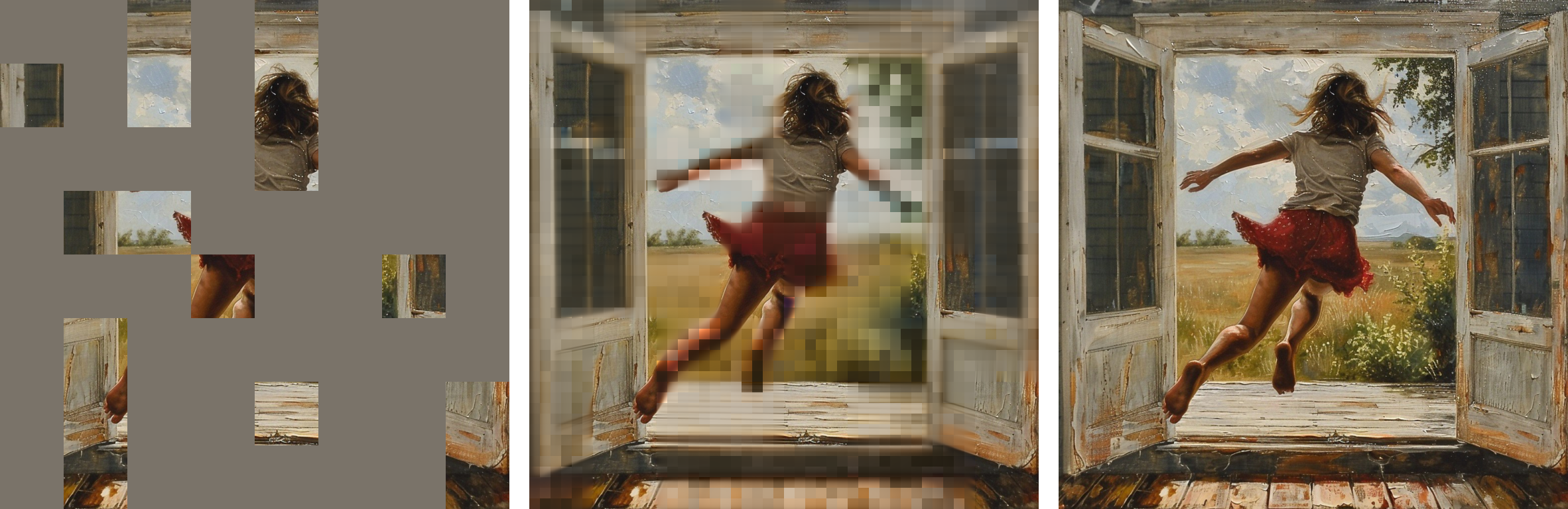}
        \caption{model learns object co-occurrence patterns (left and right doors)}
        \label{fig:mae_teaser_1}
    \end{subfigure}
    \hfill
    \begin{subfigure}{0.494\linewidth}
        \centering
        \includegraphics[width=\linewidth]{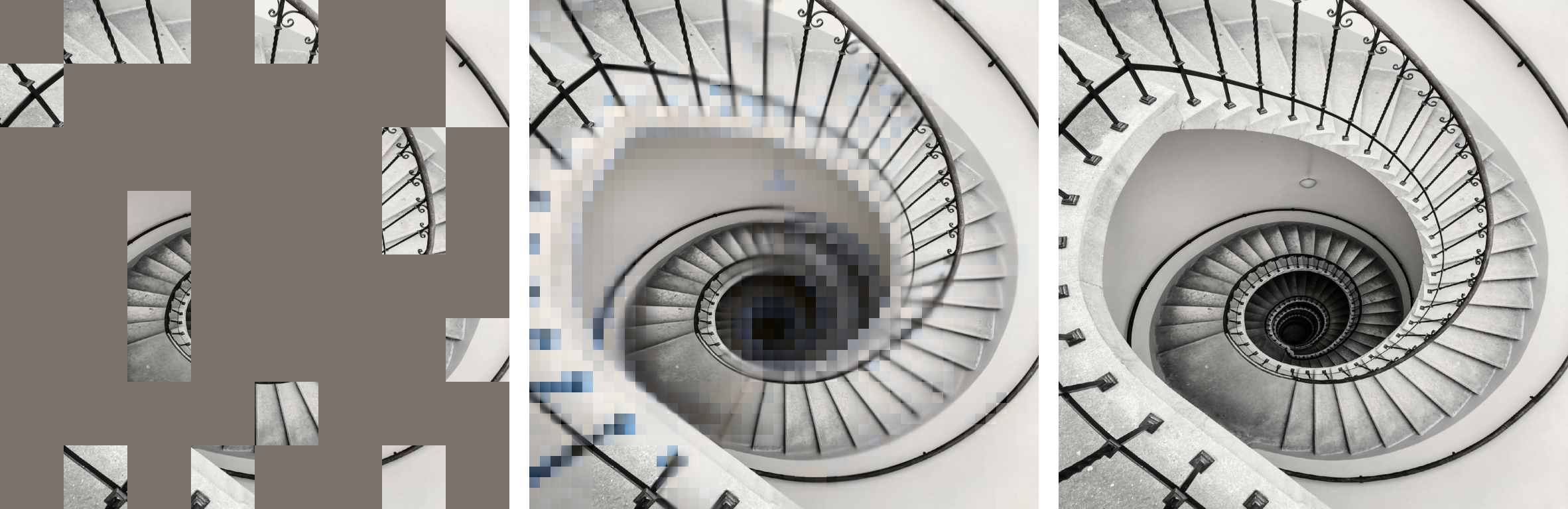}
        \caption{model infers hidden camera pose and 3D spatial layout}
        \label{fig:mae_teaser_2}
    \end{subfigure}
    \begin{subfigure}{0.494\linewidth}
    \vspace{1mm}
        \centering
        \includegraphics[width=\linewidth]{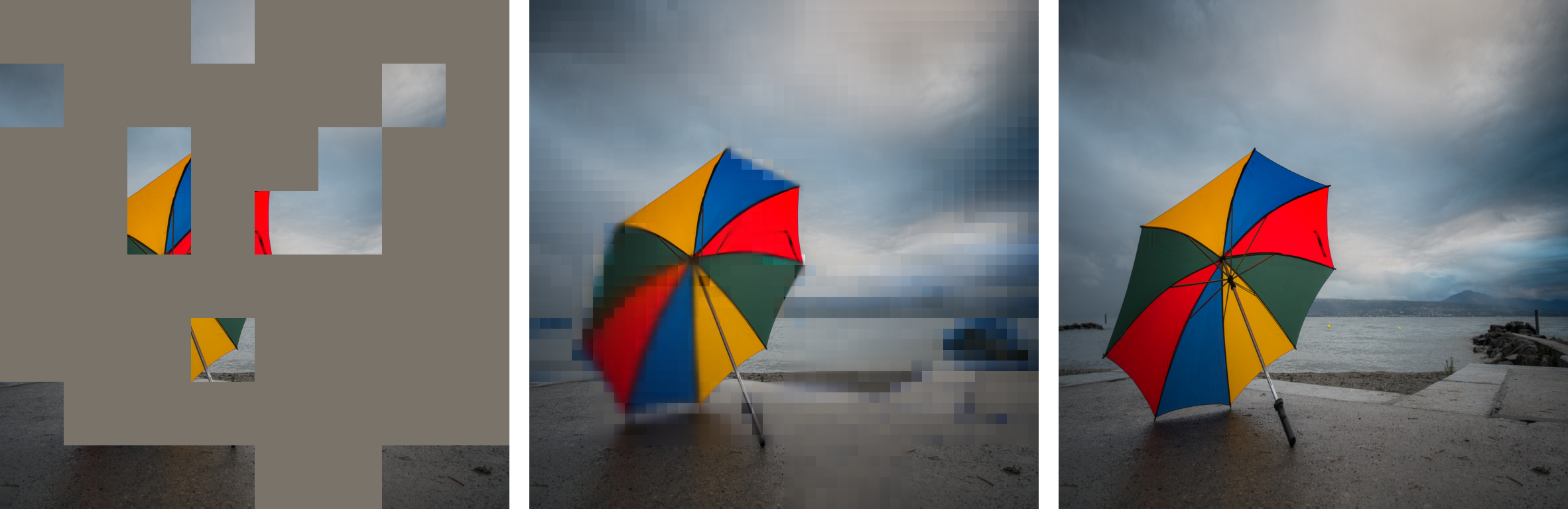}
        \caption{model reasons about symmetric color patterns (left red color is masked)}
        \label{fig:mae_teaser_3}
    \end{subfigure}
    \hfill
    \begin{subfigure}{0.494\linewidth}
        \centering
        \includegraphics[width=\linewidth]{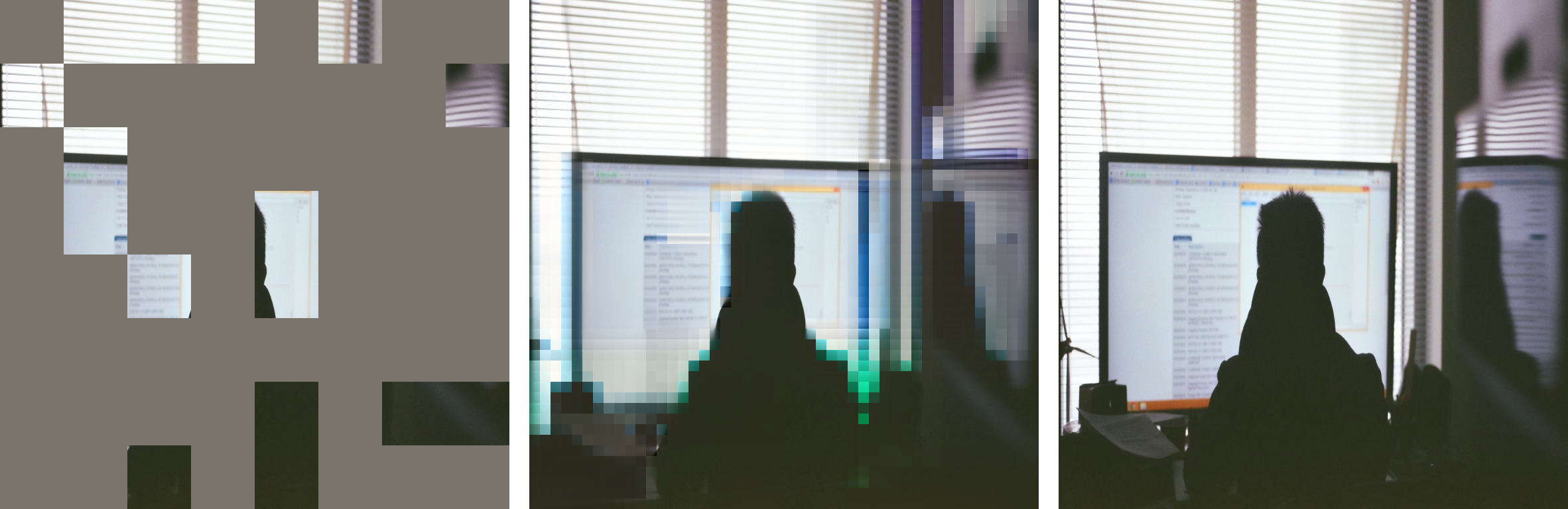}
        \caption{model understands reflection and predicts the mirrored person}
        \label{fig:mae_teaser_4}
    \end{subfigure}
    \caption{
    Pixel supervision compels the model to \textit{compress} and \textit{re-organize} visual knowledge across all levels. To accurately predict pixels, the model must understand 
    geometry, texture, semantics, materials, lighting, \etc. By masking and pixel reconstruction, MAE learns these desirable 
    visual properties and even exhibits early reasoning capabilities~\cite{wiedemer2025video}.
    From left to right in each group: masked input, reconstructed image (visible patches are kept), ground truth image (unseen during training).}
    \label{fig:mae_teaser}
\end{figure*}

In this paper, we aim to push the frontier of pixel-based self-supervised learning, without relying on any objective defined in latent spaces. We suggest that pixels are ultimately the origin of the visual information through which we perceive the physical world. They inherently contain the desired information at all levels, ranging from low-level attributes (\eg, colors, textures, materials, geometry) to high-level concepts (\eg, semantics, relations, entities, events). Rather than focus on higher-level abstractions that may treat lower-level signals as ``nuance'', we pursue generic visual representations that \mbox{\textit{compress}} and \mbox{\textit{re-organize}} information across all levels (see Figure~\ref{fig:mae_teaser}).

Our study builds on the MAE paradigm~\cite{mae}, with improvements introduced on both the algorithm and data sides. Algorithmically, we observe that the original MAE design is suboptimal in the large-data, large-model regime we investigate. To bring the models into this regime, we increase the pre-training difficulty by adopting larger masking blocks, and strengthen the model's capability by deepening the AE's decoder and enlarging the set of class tokens (see Figure~\ref{fig:main}). We empirically show that these simple designs facilitate representation learning by providing a sufficiently challenging task for a capable model to solve.

On the data side, we largely close the gap between the original MAE~\cite{mae}, trained on ImageNet~\cite{imagenet}, and the DINO family~\cite{dino,dinov2}, which benefit from training on carefully curated web-scale data.
We build training data at similar scale with less manual curation, avoiding bias toward specific benchmark distributions. 
Specifically, we collect a diverse pool of 2 billion web-crawled images~\cite{metaclip} and employ a soft self-curation strategy, where each image's sampling probability is determined by its reconstruction loss.
We demonstrate that our MAE-driven algorithm significantly benefits from such diverse, large-scale data, greatly outperforming its original version trained on the limited ImageNet-1K dataset.

We evaluate our system, codenamed ``\method'', across a diverse range of vision tasks.
We observe that in tasks where preserving lower-level details is important, such as monocular depth estimation (\eg, Depth Anything~\cite{dav2}), feed-forward 3D reconstruction (\ie, MapAnything~\cite{mapanything}), semantic segmentation, \method outperforms or matches the state-of-the-art DINOv2/v3 counterparts. 
It also delivers very promising results on robot learning tasks. 
Collectively, our results suggest that pixel-space self-supervised learning can serve as a competitive alternative and a complement to latent-space approaches.

%% file: sec/2_related_work.tex
\section{Related Work}

This section reviews the evolution of \emph{supervision signals} in visual representation learning, discussing how the definition of \emph{ground truth} has progressed.

Early deep visual representation learning~\cite{alexnet, vggnet, resnet} relies on explicit human annotations to learn transferable features. In this stage, ImageNet~\cite{imagenet} class labels are typically treated as the \emph{ground truth}, providing the primary supervision signal. However, a single categorical label is insufficient to describe an image. Such a pre-training paradigm is restricted to human-defined concepts and faces scalability challenges, yielding marginal gains when transferring to downstream tasks~\cite{he2019rethinking}.

For more scalable and richer supervision, CLIP~\cite{clip} resorts to web-crawled image-text pairs, scaling the pre-training data to 400M and facilitating open-vocabulary applications~\cite{liang2023open, gao2024clip}. However, alt-text remains an imperfect proxy for \emph{ground truth}. Captions are constrained by human knowledge and linguistic expression, often providing only partial descriptions~\cite{longclip} or even irrelevant information. Recent efforts~\cite{siglip2, metaclip2} continue to refine this paradigm, but fundamental limitations persist.

Both class labels and web captions represent projections of the physical world through human cognition and natural language. Such high-level abstractions can produce promising results in the short term~\cite{siglip, metaclip}, as they are directly distilling human knowledge. Nevertheless, they will ultimately hinder the upper bound of visual intelligence due to inherent human bias. For instance, humans naturally focus on object existence and attributes, rather than complex inter-entity relationships~\cite{wang2025spatialclip}. Many visual phenomena, \eg, subtle lighting changes, intricate spatial arrangements, or abstract aesthetic qualities, are difficult or impossible to verbalize. Moreover, such paired data is not infinitely scalable and can become a new bottleneck~\cite{villalobos2024position}.

\begin{figure*}[t]
    \centering
    \includegraphics[width=\linewidth]{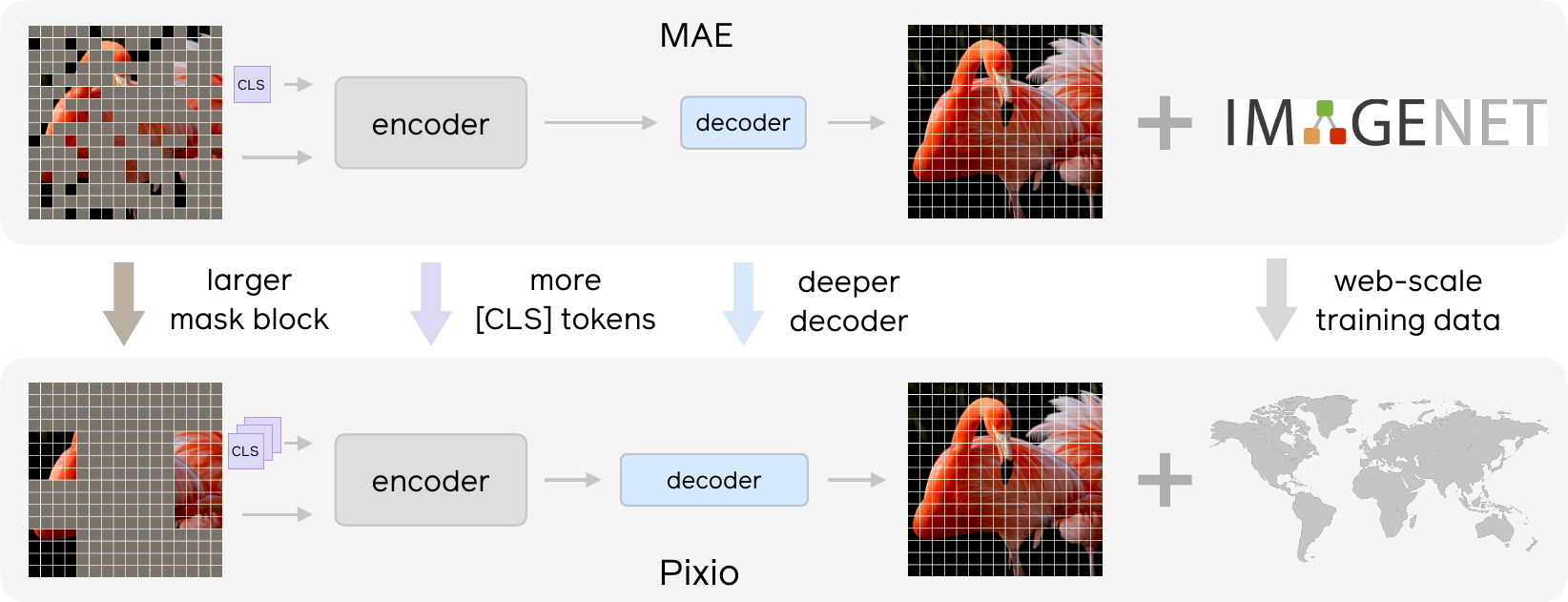}
    \caption{\method introduces four simple yet critical updates to MAE, with following motivations. \textbf{Deeper decoder:} MAE's shallow decoder lacks capacity for pixel regression, forcing the encoder to sacrifice representation quality for reconstruction. \textbf{Larger mask block:} single-patch masking causes reconstruction shortcuts and provides insufficient context. \textbf{More \texttt{[CLS]} tokens:} a single class token cannot capture diverse global properties. \textbf{Web-scale training data:} IN-1K lacks the visual diversity needed for learning transferable representations.}
    \label{fig:main}
    \vspace{2mm}
\end{figure*}

Self-supervised learning methods~\cite{instdisc, moco, mae} avoid using explicit human annotations. However, \emph{human priors} remain embedded through carefully designed pretext tasks. The inductive bias from humans implicitly serves as the \emph{ground truth} guiding models to learn the physical world. These methods typically first perform some artificial distortions on the raw input and then train the model to be invariant to them. The distortions can be simple color perturbations~\cite{simclr, mocov2}, spatial shuffling~\cite{jigsaw}, spatial masking~\cite{beit, simmim, ibot}, token removal~\cite{mae}, Gaussian noise~\cite{dae}, \etc. The supervision can be applied in either latent space~\cite{cpc, cmc, swav, data2vec, beit} or directly in pixel space~\cite{vincent2008extracting, pathak2016context, igpt, simmim, mae}. For latent-space practices, inductive bias (\eg, multi-view consistency~\cite{simclr, dino}) is necessary to prevent model collapse and drive the model to learn towards the direction as human hopes or benchmark~\cite{imagenet} prefers.

This work advocates for pixel as a promising supervision for visual representation learning. It introduces less human bias than other artificially constructed learning targets~\cite{instdisc, simclr, dino}. Humans only decide what images or videos to capture, not what features to emphasize or what invariances to enforce. Although pixel remains an imperfect representation of our complex physical world, it is fundamentally \emph{grounded in observable reality} with minimal abstraction and human intervention, especially compared to human-defined categories, human-written texts, or human-designed visual priors. We demonstrate highly capable vision models can be learned through direct pixel prediction. It offers a scalable and less biased path toward advancing visual intelligence~\cite{wiedemer2025video}.

%% file: sec/3_method.tex
\section{\method}

\method builds upon MAE with four critical modifications. So we first review MAE's core designs in Section~\ref{sec:mae}. Then, we introduce our algorithm improvements in Section~\ref{sec:pixio_framework} and pre-training data update in Section~\ref{sec:pixio_data}.

\subsection{\label{sec:mae}Preliminary of MAE}

MAE adapts BERT~\cite{bert} to visual domain. The core idea is to mask partial input signals and train the model to recover them. Three key designs distinguish MAE: 1) direct pixel supervision, 2) disentangling visible and masked tokens with an asymmetric encoder-decoder architecture, and 3) a high masking ratio to mitigate (also leverage) visual redundancy. We explain these three aspects below.

Unlike masked prediction in discrete latent space~\cite{beit, data2vec}, MAE shows pixel is simple, feasible, and more effective as direct supervision~\cite{igpt}. Notably, it does not require stabilization mechanisms such as negative samples~\cite{simclr}, stop gradient~\cite{moco}, or careful centering and sharpening~\cite{dino}.

Before MAE, masked modeling approaches~\cite{bert, vit, beit} partially replace input tokens with special \texttt{[MASK]} tokens, resulting in train-test distribution shift. Processing all tokens also imposes heavier computational burden on the encoder. MAE elegantly addresses both limitations by removing masked tokens at the encoder and appending them back at the lightweight decoder.

Compared to discrete and semantically rich text tokens, visual signals (\eg, image, video) are contiguous, low-level, and redundant~\cite{videomae, maest}. MAE demonstrates that a high masking ratio (\eg, 75\%) is necessary to avoid ground truth leakage and construct a meaningful pretext task~\cite{zhang2022mask}.

\subsection{\label{sec:pixio_framework}MAE Redesign}

We preserve the three core designs of MAE, but revisit several overlooked components that are critical for performance. Our redesigns are summarized in Figure~\ref{fig:main}.

\vspace{2mm}
\noindent
\textbf{Deeper decoder.} As shown in Figure~\ref{fig:probe_official_mae}, the best generic feature (\eg, for classification, depth estimation, semantic segmentation) in MAE does not reside in the final encoder block~\cite{alkin2024mim, bolya2025perception}. For a 32-block ViT-H encoder, optimal feature emerges as early as the 20th block.

We conjecture that the decoder lacks sufficient capacity for pixel regression. To minimize reconstruction loss, the encoder has to sacrifice some capacity (\eg, later blocks) in modeling low-level details (\ie, serve as the ``decoder''), rather than prioritizing semantic understanding as expected. Under this hypothesis, our solution becomes straightforward: \emph{increasing the decoder depth}. Simply adding more decoder blocks (while still maintaining lightweight overall overhead) yields substantial improvements (Figure~\ref{fig:decoder}).
Note that over-parameterized decoder is also not appropriate, as models may learn to prioritize memorizing visual details over learning transferable representations. Additionally, an overly powerful decoder can cause encoder laziness, with the encoder relying on the decoder for representation learning and yielding suboptimal representations.

\vspace{2mm}
\noindent
\textbf{Larger mask block.}
MAE randomly drops individual patch tokens at the encoder input. However, as illustrated in Figure~\ref{fig:main}, single-patch masking disrupts local context. More critically, the model can achieve plausible reconstruction by simply copying from nearby visible patches, without genuine visual understanding. Therefore, we \emph{mask at a larger granularity}, \eg, 4$\times$4 local patch blocks. This provides richer local context for learning and mitigates ground truth leakage. Nevertheless, excessively large masking units (\eg, 8$\times$8 patches) are not recommended, as masked regions will become unpredictable.

Large-context masking is indeed not new in masked image modeling~\cite{simmim, longmae, jepa, dinov2}. In this work, we conduct comprehensive ablation studies to provide more insights on the correlation between masking ratio, masking granularity, and downstream performance (Figure~\ref{fig:maskgrid}).

\vspace{2mm}
\noindent
\textbf{More \texttt{[CLS]} tokens.}
MAE appends a class token~\cite{bert} alongside patch tokens.
Unlike image-level contrastive learning~\cite{mocov3, dino}, this token receives no explicit loss supervision, yet works effectively in classification tasks. This token implicitly captures global information, enabling patch tokens to perform local-global interaction. 
A single class token is insufficient to capture diverse global visual properties, such as scene type, image style, object concepts, camera pose, \etc. Therefore, we \emph{append multiple class tokens}. For downstream tasks requiring a global representation, we can either average or concatenate them.

These tokens relate to ViT register tokens~\cite{darcet2023vision, capi, dinov3}, but serve different roles. Our class tokens function as global representations used directly in downstream tasks (\eg, image classification, robot learning), rather than being discarded during evaluation~\cite{darcet2023vision}.

\subsection{\label{sec:pixio_data}Web-Scale Data and Curation}

Earlier visual self-supervised learning works were mostly trained on small-scale IN-1K. Later, DINOv2~\cite{dinov2} demonstrates large-scale data with diverse concepts~\cite{vo2024automatic} is essential for learning robust and transferrable representations. However, DINOv2 and DINOv3~\cite{dinov3} employ excessive benchmark-centric data curation. For example, they use benchmark images as queries to retrieve similar training images from a large pool. They also directly inject benchmark training images (\eg, IN-1K, Mapillary~\cite{mapillary}) with excessive repetitive sampling (can be 100$\times$).

Undoubtedly, such careful curation can yield strong benchmark results in the short run. However, it may cause the model to be fragile to future unknown scenarios with different data distributions. We thus advocate pre-training on large-scale, conceptually diverse, and minimally-curated data to avoid benchmark bias.

We first follow MetaCLIP~\cite{metaclip} to collect 2 billion web-crawled images, covering substantially more diverse scenes than curated benchmarks~\cite{imagenet, yfcc}. However, the raw distribution is dominated by product images and text-heavy images (\eg, documents)~\cite{webssl}, which should not be over-emphasized in generic representation learning. We thus apply two 
complementary curation strategies.
First, we employ loss-based soft sampling. We pre-compute reconstruction loss $l$ for 
each image using a \method model trained on the raw data. We then probabilistically sample training images: image $i$ is accepted if $l_i \ge u$, where 
$u \sim \mathcal{U}(0,1)$. This strategy downsamples easy-to-reconstruct images (\eg, product images), while highlighting challenging, visually rich content. Second, we filter 
images with low histogram entropy in colors to reduce text-heavy images, which may exhibit high reconstruction loss but limited scene diversity. 
Together, these two strategies preserve rich and diverse visual content with minimal curation bias.

%% file: sec/4_exp.tex
\input{table/mde}

\input{table/depth_anything}

\section{Experiments}

Earlier self-supervised learning works~\cite{mae, dino} primarily focused on image classification tasks. Such scenarios are now more appropriate to be addressed by open-vocabulary classifiers~\cite{clip, metaclip} or MLLMs~\cite{gpt4o}. Therefore, We primarily evaluate tasks that require dense visual representations~\cite{dinov2}, such as depth estimation, 3D reconstruction, and segmentation, which remain challenging to MLLMs.

\vspace{1mm}
\noindent
\textbf{Brief introduction of our setup.}
Our largest ViT-5.4B/16 model is pre-trained on 2B curated web-crawled images for 20B seen samples over 1.3M iterations with a batch size of 16384. The input resolution is 256$\times$256. We employ a ViT decoder with 512 dimensions and 32 blocks, apply masking at 4$\times$4 patch granularity, and append 8 class tokens.
In the main paper, we mainly compare our distilled \method-H encoder (631M params) against the state-of-the-art DINOv3-H+ encoder (841M params)~\cite{dinov3}, which is the second-largest version distilled from their largest 7B model. Additional details are provided in the appendix.

\subsection{Monocular Depth Estimation}

\vspace{1mm}
\noindent
\textbf{Domain-specific monocular depth estimation.}
We freeze the pre-trained encoder and add a trainable DPT head~\cite{dpt} or linear regression head on it. The metric depth estimation model is trained and evaluated on the same domain (NYUv2~\cite{nyuv2} or KITTI~\cite{kitti}). As compared in Table~\ref{tab:mde}, our \method clearly outperforms the most capable DINOv3 model under both DPT and linear heads, \eg, reducing RMSE from 0.320 $\rightarrow$ 0.268 and improving $\delta_1$ from 93.2 $\rightarrow$ 95.5 on NYUv2. Compared to the initial MAE model, the improvement is substantial (RMSE: 0.465 $\rightarrow$ 0.268, $\delta_1$: 80.8 $\rightarrow$ 95.5), demonstrating the necessity of our modifications to both algorithm and data.

\vspace{1mm}
\noindent
\textbf{Depth Anything.}
In addition to domain-specific depth estimation, we further follow Depth Anything V2~\cite{dav2} to evalute zero-shot monocular relative depth estimation, where the model is trained on five synthetic datasets and tested on unseen distributions~\cite{sintel, diode}. As shown in Table~\ref{tab:depth_anything}, \method outperforms or matches DINOv3 on most benchmarks (\eg, NYUv2, DIODE, DA-2K), and is inferior to DINOv2/v3 on the autonomous driving benchmark KITTI. This is expected, since we do not inject abundant driving-related images as DINOv2 does, which directly uses over 1 million Mapillary images~\cite{mapillary}.

\subsection{Feed-Forward 3D Reconstruction}

Feed-forward 3D reconstruction~\cite{dust3r} requires strong visual capability in understanding spatial layout and capturing multi-view dense correspondence. We strictly follow MapAnything's~\cite{mapanything} training framework and compare the performance of different pre-trained encoders. As shown in Table~\ref{tab:mapanything}, evaluated across indoor~\cite{scannet++}, outdoor~\cite{eth3d}, and synthetic scenes~\cite{tartanair, zhang2025ufm}, \method consistently delivers better reconstruction and pose estimation results than MAE, DINOv2, and DINOv3. This demonstrates that although trained with a single view, \method exhibits stronger multi-view capability than pre-training frameworks~\cite{dinov2, dinov3} that explicitly use multiple views (\eg, 8 views in DINOv3).

\input{table/map_anything}

\subsection{Semantic Segmentation}

Semantic segmentation requires dense classification of each pixel. We evaluate on two natural image benchmarks (ADE20K~\cite{ade20k} and Pascal VOC~\cite{pascal}) and one satellite image benchmark LoveDA~\cite{loveda}. As shown in Table~\ref{tab:semseg}, \method achieves competitive results outperforming or on par with the state-of-the-art DINOv3 model, despite employing a simpler pre-training objective and requiring no benchmark-specific data curation. Note that our model uses 200M fewer parameters and is distilled from a model with 1.3B fewer parameters.

\subsection{Robot Learning}

We evaluate \method on CortexBench~\cite{cortexbench}, using four benchmarks: Adroit, DMC, MetaWorld, and Trifinger. We compare against three specialized models (VC1~\cite{cortexbench}, R3M~\cite{nair2022r3m}, Theia~\cite{shang2024theia}) and two generic models (DINOv2, DINOv3).
For fair comparison, we select the best-performed embedding configuration (either global embedding or spatial embeddings) for each model.
For models using spatial embeddings, we follow Theia~\cite{shang2024theia} to employ a three-layer CNN followed by an MLP for policy prediction.
We average the class tokens of \method as its global embedding, which we find more effective than its spatial embeddings in this scenario. It also eliminates the computational overhead of the CNN-based policy network.
As shown in Table~\ref{tab:robotics}, \method is 1.2\% better than R3M and 3.1\% better than DINOv3, indicating its strong capability in robot learning.

\input{table/semseg}

\input{table/robotics}

\begin{figure*}[t]
    \centering
    \includegraphics[width=\linewidth]{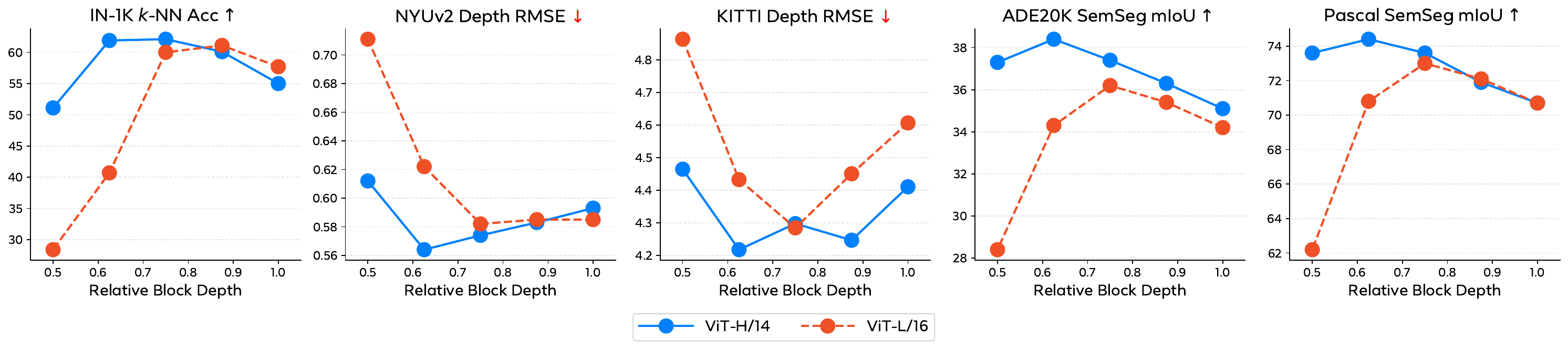}
    \caption{Probing frozen \textbf{features in different blocks} of the original MAE encoder, which is trained on ImageNet-1K. The relative block depth is computed as the ratio of the block index to the total number of blocks, for easy comparison across architectures (ViT-H: 32 blocks, ViT-L: 24 blocks). We use a linear head for both monocular depth estimation (regression) and semantic segmentation (classification).}
    \label{fig:probe_official_mae}
\end{figure*}

\begin{figure*}[t]
    \centering
    \includegraphics[width=\linewidth]{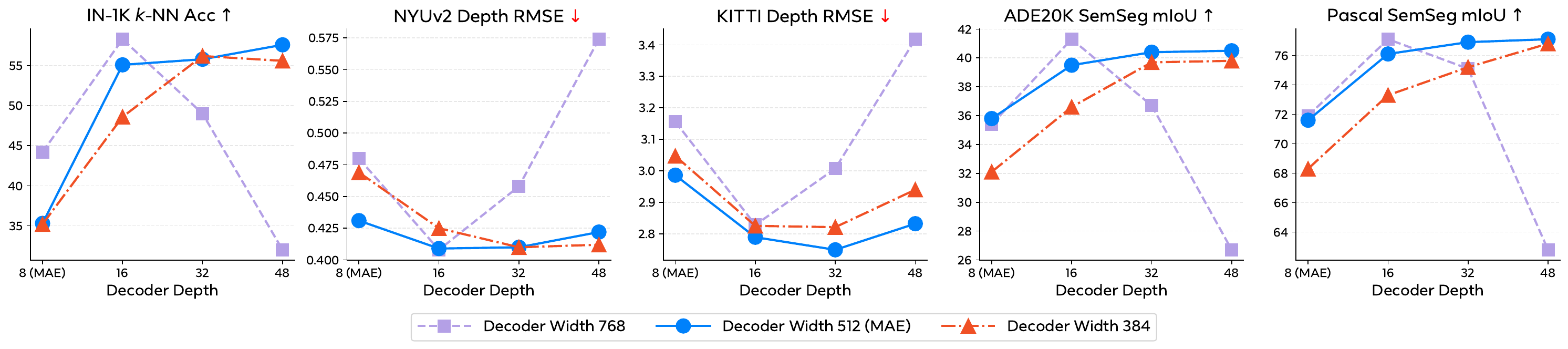}
    \caption{Ablation study of using \textbf{decoders of different depth} (\#attention blocks) or width (feature dimension) to train MAE on IN-21K. The encoder is ViT-H (1280-d $\times$ 32-blocks). Here, we use a DPT head for depth estimation and a linear head for semantic segmentation.}
    \label{fig:decoder}
\end{figure*}

\subsection{Ablation Study}

Unless otherwise specified, we train a ViT-H encoder on ImageNet-21K for 400 epochs with batch size 8192 ($\sim$5B seen images). We use IN-21K because it is publicly accessible and makes our results reproducible. For evaluation, by default, we use a DPT head for depth estimation and a linear head for semantic segmentation.

\vspace{1mm}
\noindent
\textbf{Decoder.}
As shown in Figure~\ref{fig:probe_official_mae}, the best generic feature of the MAE-H encoder is far before the last encoder block. Based on our previous analysis, increasing the decoder capacity is an intuitive approach to address this issue. In Figure~\ref{fig:decoder}, we conduct a comprehensive ablation study on the decoder's embedding dimension (width) and number of attention blocks (depth). By simply increasing the decoder depth from 8 (default) to 32, all downstream performance is tremendously enhanced, \eg, IN-1K $k$-NN accuracy 35.3 $\rightarrow$ 55.8, NYUv2 depth error 0.431 $\rightarrow$ 0.410, and ADE20K mIoU 35.8 $\rightarrow$ 40.4. Similar improvements are achieved by increasing the decoder width from 512 (default) to 768 and depth from 8 (default) to 16. However, the decoder cannot be too heavy, as it may replace the role of the encoder. Further increasing the decoder capacity (\eg, 768$\times$32) produces unsatisfactory results.

\begin{figure*}[t]
    \centering
    \includegraphics[width=\linewidth]{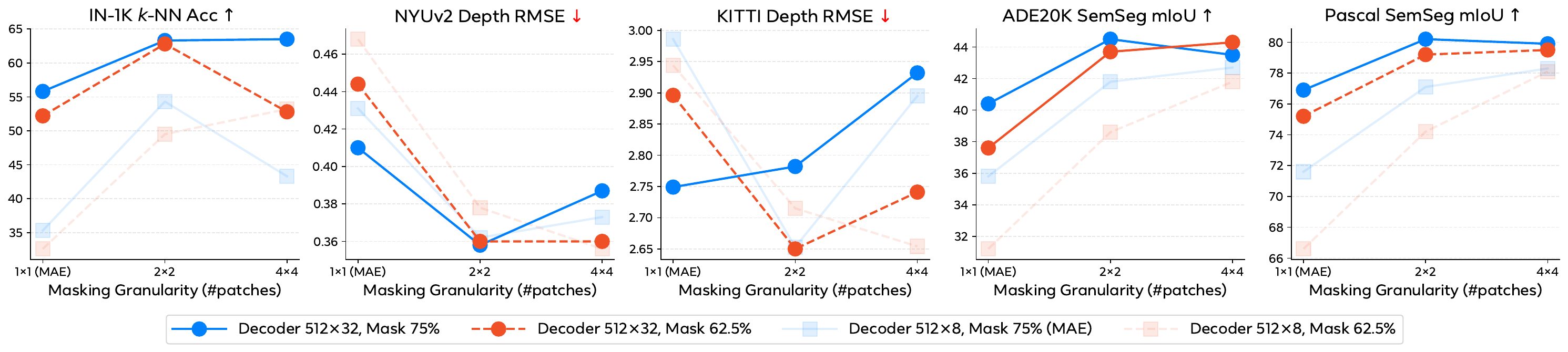}
    \caption{Ablation study on \textbf{masking granularity} (measured in \#patches). MAE uses single-patch (1$\times$1) masking granularity.}
    \label{fig:maskgrid}
\end{figure*}

\vspace{1mm}
\noindent
\textbf{Masking granularity.}
We compare performance under different masking granularity in Figure~\ref{fig:maskgrid}. Since optimal masking granularity may correlate with masking ratio, we evaluate two masking ratios: 75\% and 62.5\%. First, under MAE's default decoder configuration (512$\times$8) and masking ratio (75\%), merely changing the masking granularity from a single patch to 2$\times$2 patches improves IN-1K $k$-NN accuracy by 19.0, reduces NYUv2 depth error from 0.431 $\rightarrow$ 0.362, and improves ADE20K mIoU by 6.0. Similar trends are observed under the better 512$\times$32 decoder configuration. We also tried to further enlarge the masking context to 8$\times$8 patches, but the results are poor, as overly large masking context will cause unpredictability.

\input{table/all3modeling}

\begin{figure*}[t]
    \centering
    \includegraphics[width=\linewidth]{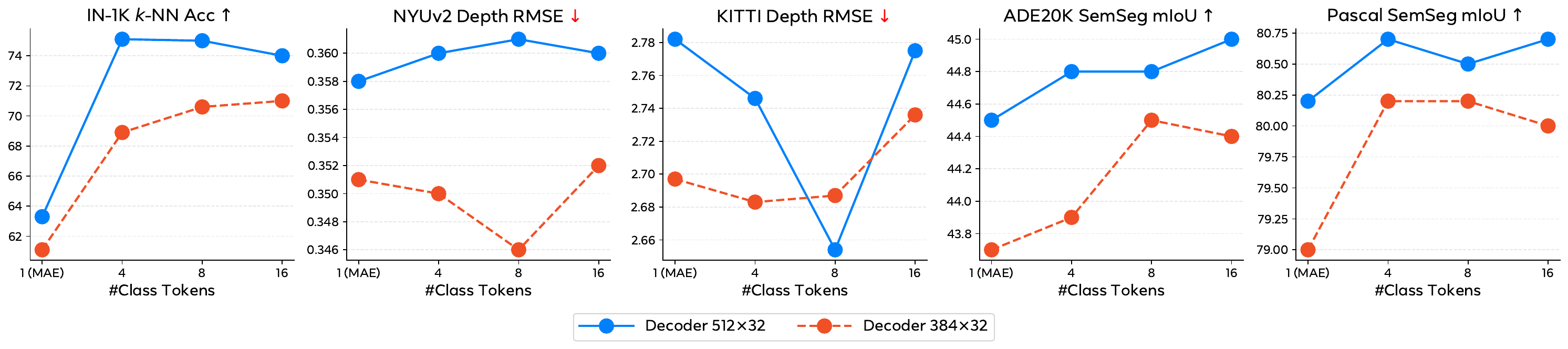}
    \caption{Ablation study on the \textbf{number of class tokens}. MAE uses a single class token.}
    \label{fig:classtoken}
\end{figure*}

\vspace{1mm}
\noindent
\textbf{Class tokens.}
MAE uses a single class token for global information. In Figure~\ref{fig:classtoken}, we examine the performance with more class tokens. The IN-1K classification is steadily improved by increasing the number of class tokens, \eg, $k$-NN accuracy 63.3 $\rightarrow$ 75.1 when increased from 1 $\rightarrow$ 4. Minor gains are also obtained on dense prediction tasks.

\input{table/data_source}

\vspace{1mm}
\noindent
\textbf{All three framework updates.} We further check the performance gain from incorporating all three proposed modifications into MAE. As shown in Table~\ref{tab:all3modeling}, \method substantially enhances MAE across image classification (37.9 $\rightarrow$ 59.5 on IN-1K), depth estimation (0.392 $\rightarrow$ 0.321 on NYUv2 $\downarrow$) and semantic segmentation (37.2 $\rightarrow$ 46.8 on ADE20K). Notably, even trained on large-scale curated 2B images, the original MAE framework fails to deliver promising results, highlighting the importance of first establishing a solid pre-training framework.

\vspace{1mm}
\noindent
\textbf{Data sources and data curation.}
Data is fundamental to self-supervised pre-training. We compare different data sources, including IN-1K, IN-21K, YFCC100M~\cite{yfcc}, and our collected 2B web images~\cite{metaclip}. All models are trained for 5B seen samples. The first three sources are delicately curated with substantial human efforts~\cite{liu2024decade}. They feature balanced yet limited visual concepts and narrow image distributions. In contrast, the web-crawled data source is diverse across all dimensions but uncurated, with dominant product images and text-heavy images (\eg, documents) that are suboptimal for generic visual pre-training. 

As shown in Table~\ref{tab:data_source}, for dense prediction tasks, IN-21K and YFCC100M substantially outperform IN-1K, demonstrating the necessity of larger-scale pre-training data. Meanwhile, although the uncurated 2B images yield inferior results compared to the highly curated IN-21K, performance is greatly improved after our simple curation procedure. We advocate for pre-training on such web-crawled sources as they are up-to-date and scalable, providing greater potential for future model scaling.

\input{table/all_model_scales}

\begin{figure}[t]
    \centering
    \includegraphics[width=1\linewidth]{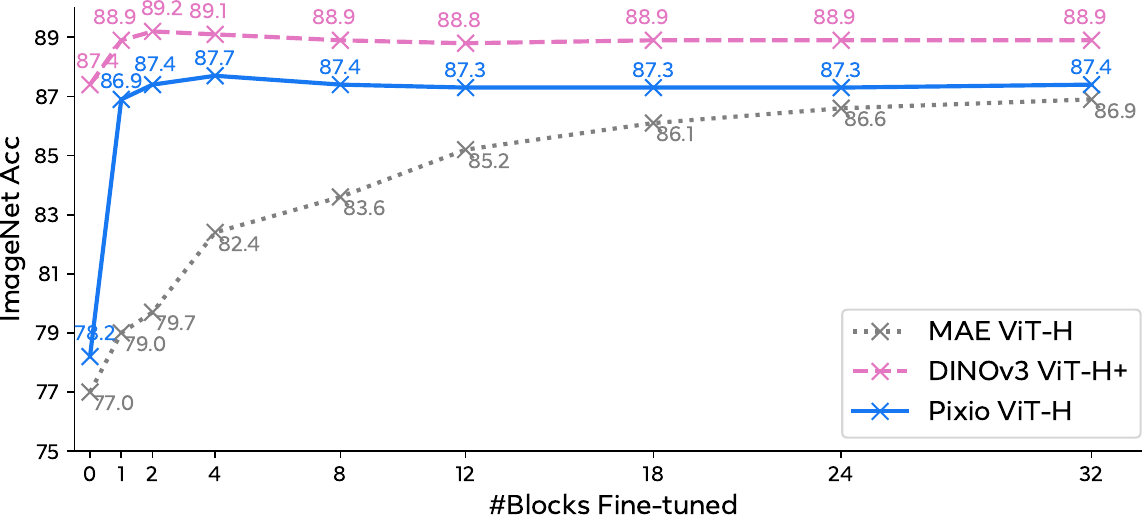}
    \caption{Fine-tuning encoder blocks for ImageNet. ``0 block'' corresponds to linear probing. \method does not intentionally involve any ImageNet data for pre-training, while DINOv3 adds ImageNet images explicitly into its training set with repetitive sampling.}
    \label{fig:in_probe_ft}
\end{figure}

\subsection{Performance of Distilled Models}

Results of our distilled more efficient student models and the largest teacher model (5B) are shown in Table~\ref{tab:all_model_scales}. Although requiring significantly fewer parameters, most students can match the performance of the teacher.

\subsection{Fine-tuning for ImageNet}

In addition to $k$-NN evaluation results on ImageNet-1K, we further examine results under fine-tuning~\cite{mae}. As compared in Figure~\ref{fig:in_probe_ft}, while \method lags behind DINOv3 in linear probing (0 block), the gap narrows substantially when further fine-tuning the last encoder block. We want to emphasize that, as a generic visual encoder, \method does not intentionally involve any ImageNet data for pre-training, while DINOv3 explicitly incorporates ImageNet-1K images (1.3M) into its training set, which constitute 10\% of its total training data (1689M) due to repetitive sampling.

%% file: table/mde.tex
\begin{table*}[t]
    \centering
    \begin{tabular}{lrrcccccccc}
    \toprule

    \multirow{3}{*}{\vspace{-3mm}Method} & \multirow{3}{*}{\vspace{-3mm}ViT} & \multirow{3}{*}{\vspace{-3mm}\#Params} & \multicolumn{4}{c}{DPT Head} & \multicolumn{4}{c}{Linear Regression Head} \\
    
    \cmidrule(lr){4-7}\cmidrule(lr){8-11}
    
    & & & \multicolumn{2}{c}{NYUv2} & \multicolumn{2}{c}{KITTI} & \multicolumn{2}{c}{NYUv2} & \multicolumn{2}{c}{KITTI} \\

    \cmidrule(lr){4-5}\cmidrule(lr){6-7}\cmidrule(lr){8-9}\cmidrule(lr){10-11}

    & & & RMSE $\downarrow$ & $\delta_1$ $\uparrow$ & RMSE $\downarrow$ & $\delta_1$ $\uparrow$ & RMSE $\downarrow$ & $\delta_1$ $\uparrow$ & RMSE $\downarrow$ & $\delta_1$ $\uparrow$ \\
    
    \midrule

    MAE & H/14 & 631M & 0.465 & 80.8 & 2.740 & 90.9 & 0.595 & 70.3 & 4.419 & 79.4 \\
    
    DINOv2 & L/14 & 304M & 0.384 & 88.0 & 2.510 & 94.7 & 0.599 & 72.0 & 4.928 & 73.7 \\
    
    DINOv2 & g/14 & 1137M & 0.355 & 90.1 & 2.424 & 94.6 & 0.560 & 75.3 & 4.425 & 78.1 \\
    
    \multirow{1}{*}{DINOv3} & H+/16 & 841M & 0.320 & 93.2 & 2.386 & 95.6 & 0.559 & 76.3 & 4.944 & 73.2 \\
    
    \midrule
    
    \multirow{1}{*}{\method} & H/16 & 631M & \textbf{0.268} & \textbf{95.5} & \textbf{2.210} & \textbf{96.7} & \textbf{0.366} & \textbf{90.8} & \textbf{3.494} & \textbf{90.3} \\
    
    \bottomrule
    \end{tabular}
    \caption{Domain-specific \textbf{monocular metric depth estimation} with frozen encoder and a trainable DPT (Dense Prediction Transformer) head or a linear \emph{regression} head.}
    \label{tab:mde}
\end{table*}

%% file: table/depth_anything.tex
\begin{table*}[t]
    \centering
    \begin{tabular}{lrrccccccccccc}
    \toprule

    \multirow{2}{*}{\vspace{-1mm}Method} & \multirow{2}{*}{\vspace{-1mm}ViT} & \multirow{2}{*}{\vspace{-1mm}\#Params} & \multicolumn{2}{c}{NYUv2} & \multicolumn{2}{c}{KITTI} & \multicolumn{2}{c}{DIODE-In} & \multicolumn{2}{c}{DIODE-Out} & \multicolumn{2}{c}{Sintel} & DA-2K \\
    
    \cmidrule(lr){4-5}\cmidrule(lr){6-7}\cmidrule(lr){8-9}\cmidrule(lr){10-11}\cmidrule(lr){12-13}\cmidrule(lr){14-14}
    
    & & & rel $\downarrow$ & $\delta_1$ $\uparrow$ & rel $\downarrow$ & $\delta_1$ $\uparrow$ & rel $\downarrow$ & $\delta_1$ $\uparrow$ & rel $\downarrow$ & $\delta_1$ $\uparrow$ & rel $\downarrow$ & $\delta_1$ $\uparrow$ & acc $\uparrow$ \\
    
    \midrule

    MAE & H/14 & 631M & 0.054 & 97.0 & 0.090 & 91.8 & 0.047 & 97.0 & 0.087 & 92.7 & 0.650 & 71.2 & 94.1 \\
    
    DINOv2 & L/14 & 304M & 0.050 & 97.6 & 0.077 & 94.0 & 0.046 & 97.6 & 0.088 & 92.8 & 0.505 & 73.4 & 95.5 \\

    DINOv2 & g/14 & 1137M & 0.046 & 97.9 & \textbf{0.074} & \textbf{94.6} & 0.041 & 98.1 & 0.085 & 93.0 & 0.450 & 75.3 & 96.9  \\
    
    \multirow{1}{*}{DINOv3} & H+/16 & 841M & 0.044 & \textbf{98.2} & 0.081 & 94.0 & 0.038 & 98.4 & 0.081 & 93.4 & \textbf{0.393} & 75.6 & \textbf{97.5} \\
    
    \midrule

    \method & H/16 & 631M & \textbf{0.041} & 98.0 & 0.083 & 93.4 & \textbf{0.034} & \textbf{98.6} & \textbf{0.075} & \textbf{94.3} & 0.535 & \textbf{75.8} & 97.3 \\
    
    \bottomrule
    \end{tabular}
    \caption{\textbf{Depth Anything V2}'s scenario for zero-shot monocular relative depth estimation. The model is trained on a combination of five synthetic datasets, and then transferred to unseen datasets. The entire model (encoder + DPT head) is trainable.}
    \label{tab:depth_anything}
\end{table*}

%% file: table/map_anything.tex
\begin{table}[t]
    \centering
    \small
    \setlength\tabcolsep{1.3mm}
    \begin{tabular}{lrcccccc}
    \toprule

    \multirow{3}{*}{Method} & \multirow{3}{*}{ViT} & \multicolumn{6}{c}{ScanNet++ v2} \\
    
    \cmidrule(lr){3-8}
    
    & & Scale & \multicolumn{2}{c}{Points} & Pose & \multicolumn{2}{c}{Depth} \\
    
    & & rel $\downarrow$ & rel $\downarrow$ & $\tau$ $\uparrow$ & auc5 $\uparrow$ & rel $\downarrow$ & $\tau$ $\uparrow$ \\
    
    \midrule

    MAE & H/14 & 0.050 & 0.057 & 63.3 & 65.6 & 0.058 & 55.4 \\
    
    DINOv2$^\dag$ & L/14 & 0.041 & 0.052 & 67.6 & 73.2 & 0.052 & 60.6 \\
    
    DINOv3 & H+/16 & 0.035 & 0.051 & 69.0 & 68.5 & 0.051 & 61.2 \\
    
    \midrule

    \method & H/16 & \textbf{0.029} & \textbf{0.041} & \textbf{78.8} & \textbf{80.5} & \textbf{0.042} & \textbf{72.4} \\

    \midrule
    \midrule

    & & \multicolumn{6}{c}{ETH3D} \\

    \midrule

    MAE & H/14 & 0.161 & 0.138 & 36.1 & 20.2 & 0.100 & 19.2 \\
    
    DINOv2$^\dag$ & L/14 & 0.204 & 0.130 & 37.6 & 23.5 & 0.095 & 24.8 \\
    
    DINOv3 & H+/16 & \textbf{0.156} & 0.146 & 39.8 & 26.9 & 0.096 & 18.6 \\
    
    \midrule

    \method & H/16 & 0.160 & \textbf{0.120} & \textbf{51.3} & \textbf{37.8} & \textbf{0.080} & \textbf{34.6} \\

    \midrule
    \midrule
    
    & & \multicolumn{6}{c}{TartanAirV2-WB} \\

    \midrule

    MAE & H/14 & \textbf{0.217} & \textbf{0.178} & 28.1 & 22.0 & 0.148 & 19.7 \\
    
    DINOv2$^\dag$ & L/14 & 0.291 & 0.222 & 30.2 & 23.6 & 0.154 & 20.6 \\
    
    DINOv3 & H+/16 & 0.342 & 0.189 & 29.9 & 22.4 & 0.133 & 24.5 \\
    
    \midrule

    \method & H/16 & 0.224 & 0.185 & \textbf{34.1} & \textbf{38.4} & \textbf{0.111} & \textbf{26.7} \\
    
    \bottomrule
    \end{tabular}
    \caption{\textbf{MapAnything}~\cite{mapanything} for feed-forward 3D reconstruction at two-view input. The input is images only. All models are trained on six Apache datasets. $\dag$: official MapAnything model.}
    \label{tab:mapanything}
\end{table}

%% file: table/semseg.tex
\begin{table}[t]
    \centering
    \small
    \setlength\tabcolsep{1.3mm}
    \begin{tabular}{lcccccc}
    \toprule

    \multirow{2}{*}{\vspace{-1mm}Model} & \multicolumn{3}{c}{DPT Head} & \multicolumn{3}{c}{Linear Head} \\

    \cmidrule(lr){2-4}\cmidrule(lr){5-7}
    
    & ADE & VOC & LoveDA & ADE & VOC & LoveDA \\
    
    \midrule

    MAE-H & 37.6 & 76.0 & 50.2 & 35.2 & 70.8 & 47.6 \\
    
    DINOv2-L & 50.1 & 84.6 & 55.2 & 47.4 & 80.2 & 52.2  \\
    
    DINOv2-g & 51.5 & 85.2 & 55.0 & 49.0 & 81.8 & 51.9 \\
    
    DINOv3-H+ & 52.3 & 85.6 & \textbf{55.3} & \textbf{50.3} & 82.1 & 52.7 \\
    
    \midrule

    \multirow{1}{*}{\method-H} & \textbf{53.6} & \textbf{85.9} & 54.7 & 50.2 & \textbf{82.2} & \textbf{53.9} \\
    
    \bottomrule
    \end{tabular}
    \caption{\textbf{Semantic segmentation} with frozen encoder and trainable DPT head or linear classification head.}
    \label{tab:semseg}
\end{table}

%% file: table/robotics.tex
\begin{table}[t]
    \centering
    \small
    \setlength\tabcolsep{1.7mm}
    \begin{tabular}{llccccc}
    \toprule

    Method & Model & Adroit & DMC & MW & Trifinger & \textbf{Avg} \\
    
    \midrule
    VC1 & ViT-L & 52.0 & 81.3 & 88.3 & 67.5 & 72.3 \\
    
    R3M & RN-50 & 74.7 & 72.4 & 94.1 & 67.4 & 77.2 \\
    
    Theia & ViT-B & 60.0 & 79.9 & 87.7 & 65.8 & 73.4 \\
    
    DINOv2 & ViT-L & 64.0 & 70.4 & 90.9 & 66.9 & 73.1 \\
    
    DINOv3 & ViT-H+ & 63.3 & 78.5 & 89.3 & 70.0 & 75.3 \\
    
    \midrule
    
    \method & ViT-H & 70.7 & 77.5 & 92.8 & 72.8 & \textbf{78.4}\\
    
    \bottomrule
    \end{tabular}
    \caption{\textbf{CortexBench for robot learning}. DMC: DeepMind Control Suite. MW: MetaWorld.}
    \label{tab:robotics}
\end{table}

%% file: table/all3modeling.tex
\begin{table}[t]
    \centering
    \small
    \setlength\tabcolsep{1.6mm}
    \begin{tabular}{lcccccc}
    \toprule
    
    \multirow{2}{*}{Method} & IN-1K & NYUv2 & KITTI & ADE20K & Pascal  \\
    
    & $k$-NN $\uparrow$ & RMSE $\downarrow$ & RMSE $\downarrow$ & mIoU $\uparrow$ & mIoU $\uparrow$ \\
    
    \midrule
    
    MAE & 37.9 & 0.392 & 2.899 & 37.2 & 67.4 \\
    
    \method & \textbf{59.5} & \textbf{0.321} & \textbf{2.581} & \textbf{46.8} & \textbf{80.2} \\
    
    \bottomrule
    \end{tabular}
    \caption{Ablation study on the combination of our framework updates. Both methods are pre-trained on the curated 2B images. MAE: decoder 512$\times$8, masking at 1$\times$1 patch, 1 class token.
    \method: decoder 512$\times$32, masking at 2$\times$2 patches, 4 class tokens.}
    \label{tab:all3modeling}
    \vspace{-2mm}
\end{table}

%% file: table/data_source.tex
\begin{table}[t]
    \centering
    \small
    \setlength\tabcolsep{1.6mm}
    \begin{tabular}{lccccc}
    \toprule

    \multirow{2}{*}{Data} & \multirow{2}{*}{Curated} & IN-1K & NYUv2 & ADE20K & Pascal \\

    & & $k$-NN $\uparrow$ & RMSE $\downarrow$ & mIoU $\uparrow$ & mIoU $\uparrow$ \\
    
    \midrule
    
    IN-1K & \cmark & \textbf{77.2} & 0.395 & 42.9 & 77.1 \\
    
    IN-21K & \cmark & 75.2 & 0.360 & 44.8 & \textbf{80.7} \\
    
    YFCC & \cmark & 59.2 & 0.345 & 44.6 & 80.3 \\

    \midrule
    
    \multirow{2}{*}{Ours: 2B} & \xmark & 54.2 & 0.351 & 44.7 & 78.0 \\

    & \cmark & 59.5 & \textbf{0.321} & \textbf{46.8} & 80.2 \\
    
    \bottomrule
    \end{tabular}
    \caption{Comparison of different training data sources. IN-1K: 1.3M images. IN-21K: 13M images. YFCC: 99M images.}
    \label{tab:data_source}
    \vspace{-2mm}
\end{table}

%% file: table/all_model_scales.tex
\begin{table}[t]
    \centering
    \small
    \setlength\tabcolsep{1.5mm}
    \begin{tabular}{lcccccc}
    \toprule
    
    \multirow{2}{*}{Model} & IN-1K & NYUv2 & KITTI & ADE20K & Pascal  \\
    
    & $k$-NN $\uparrow$ & RMSE $\downarrow$ & RMSE $\downarrow$ & mIoU $\uparrow$ & mIoU $\uparrow$ \\
    
    \midrule
    
    \method-5B & 68.4 & 0.238 & 0.213 & 50.2 & 82.0 \\

    \method-1B & 66.4 & 0.247 & 0.217 & 50.8 & 82.5 \\
    
    \method-H & 63.2 & 0.268 & 0.221 & 50.2 & 82.2 \\

    \method-L & 67.7 & 0.286 & 0.233 & 49.3 & 81.7 \\

    \method-B & 64.0 & 0.373 & 0.278 & 45.5 & 79.9 \\
    
    \bottomrule
    \end{tabular}
    \caption{Performance of various \method models. Depth estimation uses a DPT head, while semantic segmentation uses a linear head.}
    \label{tab:all_model_scales}
    \vspace{-2mm}
\end{table}

%% file: sec/6_conclusion.tex
\section{Conclusion}

This work demonstrates pixel can serve as a very promising supervision for large-scale visual pre-training, meantime enjoying simplicity, stability, and efficiency. We introduce three minimal algorithm updates to MAE and train our \method on 2B web-crawled images with a simple self-curation strategy. \method delivers results better than or comparable to the state-of-the-art DINOv3 model.

%% file: sec/X_suppl.tex
\clearpage
\setcounter{page}{1}
\maketitlesupplementary

\input{sec/5_limitation}

\section{Implementation Details}

\subsection{Pre-training}

The basic hyperparameters for our pre-training closely follow the original MAE framework, with several adaptations for large-scale training. Given our web-scale training data, we extend the training iterations from 500K to 1.3M and increase the batch size from 4,096 to 16,384. Importantly, we find that reducing the peak learning rate (2.4e-3 $\rightarrow$ 8e-4) is essential for stable convergence on less curated, more diverse web data. We also increase the input resolution from 224$\times$224 to 256$\times$256 with a patch size of 16$\times$16. Comprehensive pre-training configurations are detailed in Table~\ref{tab:implementation_details}, and the complete architecture details of our largest \method-5B model are provided in Table~\ref{tab:model_detail}.

\input{table/implementation_details}

\subsection{Distillation}

Using our pre-trained \method-5B encoder as the teacher, we distill a series of smaller, more efficient student encoders: \method-1B (1.4B parameters), \method-H (631M), \method-L (303M), and \method-B (86M). Specifically, the teacher encoder processes unmasked images while the student encoder receives either masked or the same unmasked inputs. For capable students (\eg, \method-1B, \method-H), we use masked inputs. This encourages the student to learn robust representations despite partial information. We align student features to teacher features through a lightweight MLP projection head, optimizing cosine similarity loss at both patch-token and class-token levels. The two losses are equally weighted and averaged for optimization. We use 50\% masking ratio with 4$\times$4 patch masking granularity. All students are trained for 500K iterations with batch size 8,192 and learning rate 1e-3. We use drop path~\cite{huang2016deep} 0.4 for students \method-1B and \method-H, while using drop path 0.1 for less capable students \method-L and \method-B. All other hyper-parameters remain identical to the pre-training stage.

\subsection{Downstream Evaluation}

We have open-sourced the downstream evaluation code to facilitate reproducibility. We highlight some details below.

\vspace{1mm}
\noindent
\textbf{ImageNet-1K classification.} 
For $k$-NN protocol, we follow DINO's~\cite{dino} official implementation. We report ($k$-NN) accuracy with $k=10$. For fine-tuning protocol, we follow MAE's official implementation. In both cases, images are resized to 256 pixels on the shorter side and then center-cropped to $256\times256$ for inference. For \method, we average all the class tokens to obtain the global representation.

\vspace{1mm}
\noindent
\textbf{Monocular depth estimation and semantic segmentation.} 
We evaluate under two settings: a trainable DPT head~\cite{dpt} or a linear regression/classification head, with the encoder frozen in both cases. Following DINOv2~\cite{dinov2}, we find that for certain encoders (\eg, DINOv2, MAE, \method), concatenating patch tokens with (averaged) class tokens along the channel dimension yields better performance than using patch tokens alone. We therefore report results using the optimal configuration for each encoder.
For the DPT head, we extract intermediate features evenly from four encoder stages. All models are trained for 60 epochs. To ensure fair comparison across architectures, we use training resolution 256$\times$256 for encoders with patch size 16 and 224$\times$224 for those with patch size 14, maintaining consistent effective sequence length. During inference, we apply a sliding window approach with overlap and ensemble the predictions from overlapping regions.

For Depth Anything~\cite{dav2}, MapAnything~\cite{mapanything}, and CortexBench~\cite{cortexbench}, we follow the official implementations, replacing the encoder with our pre-trained models while keeping all other components unchanged.




\section{Ablation Studies}

Limited by space in the main paper, we primarily presented ablation results through figures. Here, we provide detailed numerical results and include additional ablation configurations for completeness.

\subsection{Block-Wise Performance of MAE and \method}

Table~\ref{tab:probe_official_mae} shows the block-wise feature quality of the official MAE models. Notably, the best generic features reside far before the final encoder block. For instance, on ADE20K semantic segmentation with MAE-H, there is a substantial 3.3 mIoU performance gap between the last encoder block and the optimal intermediate block, indicating that the final layers sacrifice representation quality for reconstruction.
However, with our deeper decoder design, this issue is largely resolved. As shown in Table~\ref{tab:probe_pixio}, \method's final encoder block produces competitive features, with only a negligible 0.006 RMSE gap on NYUv2 compared to the best intermediate block. This validates our hypothesis that insufficient decoder capacity forces MAE's late encoder blocks to assume decoding responsibilities.

\input{table/probe_official_mae}

\input{table/probe_pixio}

\subsection{Decoder Design}

Beyond the decoder widths (768, 512, 384 dimensions) presented in the main paper, Table~\ref{tab:decoder} additionally reports results with an even shallower decoder (256 dimensions). The results confirm that excessively shallow decoders are suboptimal, as they lack sufficient capacity for the challenging pixel reconstruction objective.

\input{table/decoder}

\subsection{Masking Design}

Extending the analysis in the main paper, Table~\ref{tab:maskgrid} provides a comprehensive evaluation of different masking configurations (varying both masking ratio and granularity) under the 384$\times$32 decoder setting. These results further validate the importance of larger masking granularity for learning better representations.

\input{table/maskgrid}

\subsection{Number of Class Tokens}

In addition to the comparisons in the main paper, Table~\ref{tab:class_token} ablates whether class tokens should be included in the decoder input. We observe that feeding class tokens to the decoder yields slightly better performance, suggesting that allowing them to participate in reconstruction helps learn more informative global representations.

\input{table/class_token}

%% file: sec/5_limitation.tex
\section{Failure Attempts, Limitations, and Future Directions}

\subsection{Failure Attempts}

In addition to the three aforementioned modifications to MAE, we explored several other approaches that ultimately did not yield performance improvements, including but not limited to:

\begin{itemize}[topsep=1.5mm, itemsep=1.5mm]
    \item \textbf{Multi-block masking:} 
    We experimented with both inpainting (predicting center regions given surrounding context)~\cite{jepa} and outpainting variants (predicting surrounding context given center regions)~\cite{capi}. Compared to our adopted masking strategy that is based on $n\times n$ local patches, these approaches introduce additional hyper-parameter complexity, requiring careful tuning of the number of blocks, block scale ranges, and block aspect ratios. Furthermore, they constrain the diversity of masking patterns available during training. For instance, the outpainting variant consistently provides a large contiguous region of visible context, which limits the model's ability to learn long-range dependencies across spatially distant patches. Empirically, neither variant delivered performance gains over our simpler patch-block masking approach.
    
    \item \textbf{Hybrid masking ratios:}
    MAE, \method, as well as many other masked image modeling works employ a fixed masking ratio across training. However, different images may benefit from different masking ratios depending on their complexity. For simple images with high redundancy, aggressive masking ratios are necessary to create a sufficiently challenging pretext task. Conversely, for complex images with rich, non-redundant content, excessively high masking ratios can make reconstruction unpredictable, causing the model to converge to trivial solutions rather than learning meaningful representations.
    To address this, several works~\cite{kakogeorgiou2022hide, fan2023motion, shin2024self} have proposed adaptive mechanisms that dynamically determine optimal masks based on motion cues or attention maps. However, these approaches introduce additional complexity and may exhibit bias toward specific image distributions (\eg, object-centric datasets). 
    We explored a simpler alternative: hybrid masking ratios. For each training image, we randomly sample a masking ratio from a pre-defined set (\eg, [62.5\%, 75\%, 87.5\%]), allowing both simple and complex images to be trained with more appropriate difficulty levels. While this design seems conceptually reasonable, we did not observe clear improvements.
    
    \item \textbf{Hybrid masking granularity:}
    As shown in Table~\ref{tab:maskgrid}, we observe that different downstream tasks benefit from different masking granularity during pre-training. Mid-level, geometry-focused tasks (\eg, depth estimation) perform better with finer masking granularity (\eg, 2$\times$2 patch blocks), while high-level, semantics-oriented tasks (\eg, semantic segmentation) favor coarser masking granularity (\eg, 4$\times$4 patch blocks). This is expected, as easier, smaller masking units encourage the model to capture fine-grained spatial relationships, whereas harder, larger masking units promote learning of broader semantic context.
    Given these complementary benefits, a natural strategy is to employ hybrid masking granularity during training. We randomly vary the masking block size across batches to help the model adapt to different contextual scales and develop multi-level visual understanding. However, despite extensive attempts, we found that using a single, fixed masking granularity throughout training consistently yields the best performance.
    
    \item \textbf{Koleo loss on class
    tokens:} 
    DINOv2~\cite{dinov2} employs Koleo loss~\cite{koleo} to enforce uniformity in the feature distribution across samples, encouraging informative representations. However, this constitutes a strong manually-imposed inductive bias, as semantically similar samples may naturally have similar representations and should not be artificially repelled. We explored an alternative application: applying Koleo loss to our multiple class tokens rather than to individual samples. The motivation is to encourage each class token to capture distinct aspects of the image (\eg, semantics, scene layout, lighting, style), thereby promoting functional specialization among tokens. In preliminary experiments, this regularization yielded minor improvements on dense prediction tasks. However, it severely degraded ImageNet classification performance with these class tokens. More critically, we observed training instability even with very small loss weights (\eg, 0.1). Given these issues, we ultimately excluded this regularization from our framework.
    
    \item \textbf{Cross-attention in decoder:} 
    We observed that in both the original MAE and our \method, reconstructed masked regions exhibit higher visual quality than reconstructed visible regions. This occurs because the reconstruction loss is only computed on masked tokens, leading to optimization bias toward these tokens. This phenomenon also implies that the appended learnable \texttt{[MASK]} tokens and the encoder-extracted visible tokens reside in different feature spaces.
    Given this observation, we hypothesized that employing cross-attention between visible and masked tokens~\cite{crossmae}, rather than full self-attention, might better model their distinct representations while facilitating information transfer. Although this modification provided marginal computational speedup by reducing the attention complexity, it did not yield improvements in downstream performance. We therefore retained the standard self-attention mechanism in our final design.

    \item \textbf{Predicting both masked and visible patches.}
    Following the aforementioned observation of misaligned feature spaces between mask tokens and visible tokens, we explored applying reconstruction loss to both masked and visible patches. In practice, this substantially degraded model performance across downstream tasks. This finding demonstrates that plain autoencoding on all image patches is suboptimal for learning transferable visual representations, and that the asymmetric reconstruction objective is crucial to MAE's effectiveness.
    
    \item \textbf{Predicting partial masked patches:} 
    MAE reconstructs all masked patches at the decoder. However, masked patches themselves contain redundancy. Neighboring masked patches often have similar content. Also considering the increased computational cost of our deeper decoder, we attempted to reconstruct only a randomly sampled subset of masked patches~\cite{crossmae}, thereby reducing decoder overhead while maintaining the pretext task. However, while this provided marginal training speedup, it consistently degraded downstream performance.
    
    \item \textbf{Feeding multi-stage features to the decoder:}
    MAE uses only the encoder's final block output for decoding. Since pixel reconstruction requires both high-level semantic understanding and low-level textural details, relying solely on the final features may place excessive burden on the last encoder block, potentially compromising its ability to learn high-level abstractions. Ideally, different encoder stages should naturally capture different levels of visual information.
    Motivated by this, we extracted intermediate features from four encoder stages, concatenated them along the channel dimension, and fed this fused representation to the decoder. Our hypothesis was that combining early-stage and late-stage features would enable natural complementarity under pixel-level supervision. However, improvements were marginal and inconsistent across tasks. Therefore, we retained the simpler single-stage design in our final framework.
    
\end{itemize}

In summary, our three presented modifications (\ie, deeper decoder, larger masking blocks, and additional class tokens) represent minimal yet critical improvements to MAE. We highly value such simplicity in design. While some above explored alternatives may indeed be viable, we were unable to identify optimal configurations that consistently improved performance. We hope these empirical insights will inform future research in masked image modeling.
\vspace{2mm}

In addition to pre-training framework, we have also tried other data curation strategies, including but not limited to:

\begin{itemize}[topsep=1.5mm, itemsep=1.5mm]
    \item \textbf{Online hard example mining:} 
    Rather than pre-computing image difficulty using a pre-trained MAE model on uncurated data, we explored selecting informative samples dynamically during training. Specifically, at each training iteration, we performed a forward pass on a batch of $N$ candidate images and computed reconstruction loss for each. We then backpropagated only through the $k$ images with the highest reconstruction loss, where $k = \alpha \cdot N$ and $\alpha \in (0, 1]$ controls the selection ratio.
    However, this approach proved problematic in practice. Early in training, when the model has not yet learned meaningful representations, the loss-based difficulty estimation is unreliable and noisy. This instability can lead to suboptimal convergence, as the training distribution shifts unpredictably. Therefore, we adopted the offline pre-computation strategy instead, which provides more stable difficulty estimates.
    
    \item \textbf{Canny edge density as a proxy for sample difficulty:} 
    In addition to reconstruction loss, we explored using Canny edge density~\cite{canny}, which is measured as the proportion of edge pixels detected in an image, as a heuristic proxy for image complexity. The intuition is that images with richer edge structures may contain more informative visual content. However, we found that such hand-crafted edge detectors are overly sensitive to low-level patterns and repetitive textures (\eg, grass, fabric patterns, brick walls), which produce high edge responses but offer limited semantic diversity.

\end{itemize}

\subsection{Limitations and Future Directions}

As the name conveys, MAE's core principles are \emph{masking} and \emph{autoencoding}. This work advocates for \emph{pixel supervision} in visual pre-training. Pixel supervision shares philosophical similarities with autoencoding: both use models to compress and reconstruct input signals. We believe pixels provide the most comprehensive and unbiased supervision for pre-training, capturing rich visual information with minimal human intervention.

However, we recognize fundamental limitations in applying \emph{masking to static images}. On the positive side, masking constructs a meaningful pretext task. High masking ratios encourage models to learn both high-level semantics and low-level details. Nevertheless, random masking remains an \emph{artificial distortion}, which introduces undesirable biases.
In practice, masking presents unavoidable trade-offs. Low masking ratios cause ground truth leakage, making reconstruction trivial. High masking ratios provide insufficient context for learning and create distribution shift between training and inference. Critically, masked image modeling never exposes the model to natural, complete images during training. Despite these limitations, masking (or other artificial distortions) appears necessary for image-based pre-training.

The fundamental reason why so many artificial distortions and human inductive biases are necessary is, \emph{static images} have inherent limitations as a medium for learning visual intelligence. Images are not the natural format in which visual information exists in the physical world. Humans do not learn from isolated snapshots. Instead, we learn through continuous temporal experiences in a causal manner, observing how the world evolves over time.
From this perspective, \emph{video} deserves greater emphasis, particularly long videos that capture the natural progression of events and their causal relationships. Videos offer a crucial advantage: the temporal dimension enables natural predictive objectives without artificial masking. Models can predict future frames from current observations—a task grounded in the causal structure of the physical world. This eliminates the need for artificial spatial masking~\cite{videomae} or noise injection~\cite{dae}. 

This work serves as a pioneering validation that pixel supervision \emph{alone} can produce strong visual representations competitive with more complex pre-training paradigms. Looking forward, we will scale this supervision approach to web-scale video data. By leveraging the temporal richness of videos and natural predictive objectives, we aim to develop more powerful and less biased visual foundation models for both videos and images.

%% file: table/implementation_details.tex
\begin{table}[h]
    \centering
    \begin{minipage}{0.55\textwidth}
        \begin{tabular}{l@{\hspace{1cm}}l}
        \toprule
        config & value \\
        \midrule
    
        data & 2B web-crawled images \\
        
        iterations & 1,284,000 \\
        
        batch size & 16,384 \\
    
        input resolution & 256$\times$256 \\
    
        precision & bfloat16 \\
        
        optimizer & AdamW \\
        
        learning rate & 8e-4 \\
        
        learning rate schedule & cosine decay \\
    
        warmup steps & 128,400 \\
        
        weight decay & 0.05 \\
    
        optimizer momentum & $\beta_1, \beta_2$ = 0.9, 0.95 \\
    
        data augmentation & RandomResizedCrop \\
    
        crop scale & 0.2-1.0 \\
    
        crop ratio & 0.75-1.33 \\
    
        drop path & 0.4 \\
        
        masking ratio & 75\% \\
        
        masking granularity & 4$\times$4 patches \\
        
        \bottomrule
        
        \end{tabular}

    \caption{Pre-training details.}
    \label{tab:implementation_details}
    \end{minipage}
    \hfill
    \begin{minipage}{0.4\textwidth}
        \centering
        \renewcommand{\arraystretch}{0.985}
        \begin{tabular}{l@{\hspace{1.5cm}}l}
        \toprule

        config & value \\

        \midrule

        \multicolumn{2}{c}{encoder} \\

        \midrule
        \#params & 5.4B \\

        patch size & 16$\times$16 \\
        
        \#blocks & 48 \\

        embedding dimension & 3072 \\
        
        hidden dimension & 12288 \\

        attention heads & 32 \\
        
        positional embedding & learnable \\
        
        \#class tokens & 8 \\

        \midrule

        \multicolumn{2}{c}{decoder} \\

        \midrule

        \#params & 103M \\
        
        \#blocks & 32 \\

        embedding dimension & 512 \\

        hidden dimension & 2048 \\
        
        attention heads & 16 \\

        positional embedding & learnable \\
        
        \bottomrule
        \end{tabular}
        \caption{Teacher model details.}
        \label{tab:model_detail}
    \end{minipage}
\end{table}

%% file: table/probe_official_mae.tex
\begin{table}[h]
    \centering
    \begin{tabular}{rcccccc}
    \toprule
    ViT & Block Index & IN-1K KNN $\uparrow$ & NYUv2 RMSE $\downarrow$ & KITTI RMSE $\downarrow$ & ADE20K mIoU $\uparrow$ & Pascal mIoU $\uparrow$ \\

    \midrule

    \multirow{5}{*}{H/14} & 32 & 55.0 & 0.593 & 4.411 & 35.1 & 70.7 \\

    & 28 & 60.1 & 0.583 & 4.247 & 36.3 & 71.9 \\

    & 24 & \textbf{62.1} & 0.574 & 4.298 & 37.4 & 73.6 \\

    & 20 & 61.9 & \textbf{0.564} & \textbf{4.218} & \textbf{38.4} & \textbf{74.4} \\

    & 16 & 51.1 & 0.612 & 4.465 & 37.3 & 73.6 \\
    
    \midrule

    \multirow{5}{*}{L/16} & 24 & 57.7 & 0.585 & 4.607 & 34.2 & 70.7 \\
    
    & 21 & \textbf{61.1} & 0.585 & 4.451 & 35.4 & 72.1 \\

    & 18 & 60.0 & \textbf{0.582} & \textbf{4.285} & \textbf{36.2} & \textbf{73.0} \\
    
    & 15 & 40.7 & 0.622 & 4.433 & 34.3 & 70.8 \\

    & 12 & 28.4 & 0.711 & 4.864 & 28.4 & 62.2 \\
    
    \bottomrule
    \end{tabular}
    \caption{Probing officially released MAE-H/14 (1280$\times$32) and MAE-L/16 (1024 $\times$24) encoders~\cite{mae}, which are trained on ImageNet-1K. Decoder: 512$\times$8. We use a linear head for both monocular depth estimation (regression) and semantic segmentation (classification).}
    \label{tab:probe_official_mae}
\end{table}

%% file: table/probe_pixio.tex
\begin{table}[h]
    \centering
    \begin{tabular}{cccccc}
    \toprule
    Block Index & IN-1K KNN $\uparrow$ & NYUv2 RMSE $\downarrow$ & KITTI RMSE $\downarrow$ & ADE20K mIoU $\uparrow$ & Pascal mIoU $\uparrow$ \\

    \midrule

    48 & 68.4 & 0.360 & 3.603 & 50.2 & 82.0 \\

    42 & 69.7 & \textbf{0.354} & 3.583 & 50.3 & \textbf{82.4} \\

    36 & 70.4 & 0.361 & \textbf{3.570} & \textbf{50.7} & 81.9 \\

    30 & \textbf{70.8} & 0.370 & 3.579 & 50.3 & 82.0 \\

    24 & 70.2 & 0.390 & 3.575 & 49.8 & 81.6 \\
    
    \bottomrule
    \end{tabular}
    \caption{Probing our \method-5.4B encoder (3072$\times$48), which is trained from scratch on our curated 2B images. Decoder: 512$\times$32. We use a linear head for both monocular depth estimation (regression) and semantic segmentation (classification).}
    \label{tab:probe_pixio}
\end{table}

%% file: table/decoder.tex
\begin{table}[h]
    \centering
    \begin{tabular}{ccccccc}
    \toprule
    \multicolumn{2}{c}{Decoder} & \multirow{2}{*}{IN-1K KNN $\uparrow$} & \multirow{2}{*}{NYUv2 RMSE $\downarrow$} & \multirow{2}{*}{KITTI RMSE $\downarrow$} & \multirow{2}{*}{ADE20K mIoU $\uparrow$} & \multirow{2}{*}{Pascal mIoU $\uparrow$} \\

    \cmidrule(lr){1-2}
    
    Width & Depth \\
    
    \midrule
    
    \multirow{4}{*}{768} & 8 & 44.2 & 0.480 & 3.156 & 35.4 & 71.9 \\

    & 16 & 58.3 & 0.408 & 2.828 & 41.3 & 77.1 \\

    & 32 & 49.0 & 0.458 & 3.007 & 36.7 & 75.1 \\

    & 48 & 32.0 & 0.574 & 3.418 & 26.7 & 62.8 \\

    \midrule

    \multirow{4}{*}{512} & 8 (MAE) & 35.3 & 0.431 & 2.986 & 35.8 & 71.6 \\

    & 16 & 55.1 & 0.409 & 2.789 & 39.5 & 76.1 \\

    & 32 & 55.8 & 0.410 & 2.749 & 40.4 & 76.9 \\

    & 48 & 57.6 & 0.422 & 2.832 & 40.5 & 77.1 \\

    \midrule

    \multirow{4}{*}{384} & 8 & 35.2 & 0.469 & 3.047 & 32.1 & 68.3 \\

    & 16 & 48.6 & 0.425 & 2.825 & 36.6 & 73.3 \\

    & 32 & 56.2 & 0.410 & 2.821 & 39.7 & 75.2 \\

    & 48 & 55.6 & 0.412 & 2.940 & 39.8 & 76.8 \\

    \midrule

    \multirow{4}{*}{256} & 8 & 32.6 & 0.499 & 2.995 & 29.1 & 64.8 \\

    & 16 & 38.6 & 0.473 & 3.001 & 32.4 & 68.5 \\

    & 32 & 47.2 & 0.451 & 2.923 & 35.7 & 71.2 \\

    & 48 & 43.7 & 0.437 & 2.898 & 37.4 & 74.4 \\
    
    \bottomrule
    \end{tabular}
    \caption{Ablation study on the decoder on a ViT-H encoder (1280-dim$\times$32-blocks). Here we mask at a single patch and use 1 class token.}
    \label{tab:decoder}
\end{table}

%% file: table/maskgrid.tex
\begin{table}[h]
    \centering
    \setlength\tabcolsep{1.4mm}
    \begin{tabular}{cccccccc}
    \toprule
    \multirow{2}{*}{Decoder} & \multicolumn{2}{c}{Masking} & \multirow{2}{*}{IN-1K KNN $\uparrow$} & \multirow{2}{*}{NYUv2 RMSE $\downarrow$} & \multirow{2}{*}{KITTI RMSE $\downarrow$} & \multirow{2}{*}{ADE20K mIoU $\uparrow$} & \multirow{2}{*}{Pascal mIoU $\uparrow$} \\
    
    \cmidrule(lr){2-3}
    
    & Ratio & Granularity \\
    
    \midrule

    \multirow{6}{*}{512$\times$8} & \multirow{3}{*}{75\%} & 1$\times$1 (MAE) & 35.3 & 0.431 & 2.986 & 35.8 & 71.6  \\

    &  & 2$\times$2 & 54.3 & 0.362 & 2.653 & 41.8 & 77.1 \\

    &  & 4$\times$4 & 43.3 & 0.373 & 2.895 & 42.7 & 78.3 \\

    & \multirow{3}{*}{62.5\%} & 1$\times$1  & 32.6 & 0.468 & 2.944 & 31.2 & 66.6 \\
    
    &  & 2$\times$2 & 49.5 & 0.378 & 2.715 & 38.6 & 74.2  \\
    
    &  & 4$\times$4 & 53.2 & 0.356 & 2.654 & 41.8 & 78.1 \\
    
    \midrule

    \multirow{6}{*}{512$\times$32} & \multirow{3}{*}{75\%} & 1$\times$1 & 55.8 & 0.410 & 2.749 & 40.4 & 76.9 \\
    
    &  & 2$\times$2 & 63.3 & 0.358 & 2.782 & 44.5 & 80.2 \\

    &  & 4$\times$4 & 63.5 & 0.387 & 2.932 & 43.5 & 79.9 \\

    & \multirow{3}{*}{62.5\%} & 1$\times$1 & 52.2 & 0.444 & 2.896 & 37.6 & 75.2 \\

    &  & 2$\times$2 & 62.8 & 0.360 & 2.650 & 43.7 & 79.2 \\
    
    &  & 4$\times$4 & 52.8 & 0.360 & 2.741 & 44.3 & 79.5 \\
    
    \midrule
    
    \multirow{6}{*}{384$\times$32} & \multirow{3}{*}{75\%} & 1$\times$1 & 56.2 & 0.410 & 2.821 & 39.7 & 75.2 \\
     
    &  & 2$\times$2 & 61.1 & 0.351 & 2.697 & 43.7 & 79.0 \\

    &  & 4$\times$4 & 61.8 & 0.366 & 2.909 & 44.4 & 80.2 \\

    & \multirow{3}{*}{62.5\%} & 1$\times$1 & 46.6 & 0.450 & 2.907 & 35.3 & 72.1 \\

    &  & 2$\times$2 & 57.0 & 0.359 & 2.675 & 42.4 & 78.2 \\

    &  & 4$\times$4 & 57.9 & 0.357 & 2.725 & 44.5 & 79.4 \\
    
    \bottomrule
    \end{tabular}
    \caption{Ablation study on masking ratio and masking granularity (measured in \#patches). Here we use 1 class token.}
    \label{tab:maskgrid}
\end{table}

%% file: table/class_token.tex
\begin{table}[h]
    \centering
    \setlength\tabcolsep{1.4mm}
    \begin{tabular}{cccccccc}
    \toprule
    Decoder & \#\texttt{[CLS]} & In Decoder & IN-1K KNN $\uparrow$ & NYUv2 RMSE $\downarrow$ & KITTI RMSE $\downarrow$ & ADE20K mIoU $\uparrow$ & Pascal mIoU $\uparrow$ \\

    \midrule

    \multirow{8}{*}{512$\times$32} & 1 (MAE) & \cmark & 63.3 & 0.358 & 2.782 & 44.5 & 80.2 \\

    & 4 & \cmark & 75.1 & 0.360 & 2.746 & 44.8 & 80.7 \\

    & 8 & \cmark & 75.0 & 0.361 & 2.654 & 44.8 & 80.5 \\

    & 16 & \cmark & 74.0 & 0.360 & 2.775 & 45.0 & 80.7 \\

    \cmidrule(l){2-8}

    & 1 & \xmark & 64.1 & 0.373 & 2.787 & 44.3 & 80.1 \\
    
    & 4 & \xmark & 68.9 & 0.364 & 2.663 & 44.8 & 80.2 \\

    & 8 & \xmark & 70.6 & 0.376 & 2.794 & 44.2 & 80.0 \\

    & 16 & \xmark & 71.9 & 0.373 & 2.728 & 44.2 & 80.4 \\
    
    \midrule
    
    \multirow{8}{*}{384$\times$32} & 1 (MAE) & \cmark & 61.1 & 0.351 & 2.697 & 43.7 & 79.0 \\
    
    & 4 & \cmark & 68.9 & 0.350 & 2.683 & 43.9 & 80.2 \\

    & 8 & \cmark & 70.6 & 0.346 & 2.687 & 44.5 & 80.2 \\

    & 16 & \cmark & 71.0 & 0.352 & 2.736 & 44.4 & 80.0 \\

    \cmidrule(l){2-8}

    & 1 & \xmark & 62.6 & 0.362 & 2.688 & 43.9 & 79.3 \\

    & 4 & \xmark & 66.0 & 0.369 & 2.784 & 43.3 & 79.7 \\

    & 8 & \xmark & 68.7 & 0.361 & 2.762 & 44.1 & 79.7 \\

    & 16 & \xmark & 70.6 & 0.356 & 2.715 & 44.3 & 80.0 \\
    
    \bottomrule
    \end{tabular}
    \caption{Ablation study on the number of class tokens and whether to include them in the decoder. Here we mask at 2$\times$2 patch blocks.}
    \label{tab:class_token}
\end{table}

%% file: main.bbl
\begin{thebibliography}{83}
\providecommand{\natexlab}[1]{#1}
\providecommand{\url}[1]{\texttt{#1}}
\expandafter\ifx\csname urlstyle\endcsname\relax
  \providecommand{\doi}[1]{doi: #1}\else
  \providecommand{\doi}{doi: \begingroup \urlstyle{rm}\Url}\fi

\bibitem[Alkin et~al.(2025)Alkin, Miklautz, Hochreiter, and Brandstetter]{alkin2024mim}
Benedikt Alkin, Lukas Miklautz, Sepp Hochreiter, and Johannes Brandstetter.
\newblock Mim-refiner: A contrastive learning boost from intermediate pre-trained representations.
\newblock In \emph{ICLR}, 2025.

\bibitem[Assran et~al.(2023)Assran, Duval, Misra, Bojanowski, Vincent, Rabbat, LeCun, and Ballas]{jepa}
Mahmoud Assran, Quentin Duval, Ishan Misra, Piotr Bojanowski, Pascal Vincent, Michael Rabbat, Yann LeCun, and Nicolas Ballas.
\newblock Self-supervised learning from images with a joint-embedding predictive architecture.
\newblock In \emph{CVPR}, 2023.

\bibitem[Baevski et~al.(2022)Baevski, Hsu, Xu, Babu, Gu, and Auli]{data2vec}
Alexei Baevski, Wei-Ning Hsu, Qiantong Xu, Arun Babu, Jiatao Gu, and Michael Auli.
\newblock Data2vec: A general framework for self-supervised learning in speech, vision and language.
\newblock In \emph{ICML}, 2022.

\bibitem[Bao et~al.(2022)Bao, Dong, Piao, and Wei]{beit}
Hangbo Bao, Li Dong, Songhao Piao, and Furu Wei.
\newblock Beit: Bert pre-training of image transformers.
\newblock In \emph{ICLR}, 2022.

\bibitem[Bolya et~al.(2025)Bolya, Huang, Sun, Cho, Madotto, Wei, Ma, Zhi, Rajasegaran, Rasheed, et~al.]{bolya2025perception}
Daniel Bolya, Po-Yao Huang, Peize Sun, Jang~Hyun Cho, Andrea Madotto, Chen Wei, Tengyu Ma, Jiale Zhi, Jathushan Rajasegaran, Hanoona Rasheed, et~al.
\newblock Perception encoder: The best visual embeddings are not at the output of the network.
\newblock In \emph{NeurIPS}, 2025.

\bibitem[Butler et~al.(2012)Butler, Wulff, Stanley, and Black]{sintel}
Daniel~J Butler, Jonas Wulff, Garrett~B Stanley, and Michael~J Black.
\newblock A naturalistic open source movie for optical flow evaluation.
\newblock In \emph{ECCV}, 2012.

\bibitem[Canny(2009)]{canny}
John Canny.
\newblock A computational approach to edge detection.
\newblock \emph{TPAMI}, 2009.

\bibitem[Caron et~al.(2020)Caron, Misra, Mairal, Goyal, Bojanowski, and Joulin]{swav}
Mathilde Caron, Ishan Misra, Julien Mairal, Priya Goyal, Piotr Bojanowski, and Armand Joulin.
\newblock Unsupervised learning of visual features by contrasting cluster assignments.
\newblock In \emph{NeurIPS}, 2020.

\bibitem[Caron et~al.(2021)Caron, Touvron, Misra, J{\'e}gou, Mairal, Bojanowski, and Joulin]{dino}
Mathilde Caron, Hugo Touvron, Ishan Misra, Herv{\'e} J{\'e}gou, Julien Mairal, Piotr Bojanowski, and Armand Joulin.
\newblock Emerging properties in self-supervised vision transformers.
\newblock In \emph{ICCV}, 2021.

\bibitem[Chen et~al.(2020{\natexlab{a}})Chen, Radford, Child, Wu, Jun, Luan, and Sutskever]{igpt}
Mark Chen, Alec Radford, Rewon Child, Jeffrey Wu, Heewoo Jun, David Luan, and Ilya Sutskever.
\newblock Generative pretraining from pixels.
\newblock In \emph{ICML}, 2020{\natexlab{a}}.

\bibitem[Chen et~al.(2020{\natexlab{b}})Chen, Kornblith, Norouzi, and Hinton]{simclr}
Ting Chen, Simon Kornblith, Mohammad Norouzi, and Geoffrey Hinton.
\newblock A simple framework for contrastive learning of visual representations.
\newblock In \emph{ICML}, 2020{\natexlab{b}}.

\bibitem[Chen et~al.(2020{\natexlab{c}})Chen, Fan, Girshick, and He]{mocov2}
Xinlei Chen, Haoqi Fan, Ross Girshick, and Kaiming He.
\newblock Improved baselines with momentum contrastive learning.
\newblock \emph{arXiv:2003.04297}, 2020{\natexlab{c}}.

\bibitem[Chen et~al.(2021)Chen, Xie, and He]{mocov3}
Xinlei Chen, Saining Xie, and Kaiming He.
\newblock An empirical study of training self-supervised vision transformers.
\newblock In \emph{ICCV}, 2021.

\bibitem[Chen et~al.(2025)Chen, Liu, Xie, and He]{dae}
Xinlei Chen, Zhuang Liu, Saining Xie, and Kaiming He.
\newblock Deconstructing denoising diffusion models for self-supervised learning.
\newblock In \emph{ICLR}, 2025.

\bibitem[Chuang et~al.(2025)Chuang, Li, Wang, Yeh, Lyu, Raghavendra, Glass, Huang, Weston, Zettlemoyer, et~al.]{metaclip2}
Yung-Sung Chuang, Yang Li, Dong Wang, Ching-Feng Yeh, Kehan Lyu, Ramya Raghavendra, James Glass, Lifei Huang, Jason Weston, Luke Zettlemoyer, et~al.
\newblock Meta clip 2: A worldwide scaling recipe.
\newblock In \emph{NeurIPS}, 2025.

\bibitem[Darcet et~al.(2024)Darcet, Oquab, Mairal, and Bojanowski]{darcet2023vision}
Timoth{\'e}e Darcet, Maxime Oquab, Julien Mairal, and Piotr Bojanowski.
\newblock Vision transformers need registers.
\newblock In \emph{ICLR}, 2024.

\bibitem[Darcet et~al.(2025)Darcet, Baldassarre, Oquab, Mairal, and Bojanowski]{capi}
Timoth{\'e}e Darcet, Federico Baldassarre, Maxime Oquab, Julien Mairal, and Piotr Bojanowski.
\newblock Cluster and predict latent patches for improved masked image modeling.
\newblock \emph{TMLR}, 2025.

\bibitem[Deng et~al.(2009)Deng, Dong, Socher, Li, Li, and Fei-Fei]{imagenet}
Jia Deng, Wei Dong, Richard Socher, Li-Jia Li, Kai Li, and Li Fei-Fei.
\newblock Imagenet: A large-scale hierarchical image database.
\newblock In \emph{CVPR}, 2009.

\bibitem[Devlin et~al.(2019)Devlin, Chang, Lee, and Toutanova]{bert}
Jacob Devlin, Ming-Wei Chang, Kenton Lee, and Kristina Toutanova.
\newblock Bert: Pre-training of deep bidirectional transformers for language understanding.
\newblock In \emph{NAACL}, 2019.

\bibitem[Dosovitskiy(2021)]{vit}
Alexey Dosovitskiy.
\newblock An image is worth 16x16 words: Transformers for image recognition at scale.
\newblock In \emph{ICLR}, 2021.

\bibitem[Everingham et~al.(2010)Everingham, Van~Gool, Williams, Winn, and Zisserman]{pascal}
Mark Everingham, Luc Van~Gool, Christopher~KI Williams, John Winn, and Andrew Zisserman.
\newblock The pascal visual object classes (voc) challenge.
\newblock \emph{IJCV}, 2010.

\bibitem[Fan et~al.(2023)Fan, Wang, Liao, Zhu, Bhat, Santos-Villalobos, MV, and Li]{fan2023motion}
David Fan, Jue Wang, Shuai Liao, Yi Zhu, Vimal Bhat, Hector Santos-Villalobos, Rohith MV, and Xinyu Li.
\newblock Motion-guided masking for spatiotemporal representation learning.
\newblock In \emph{ICCV}, 2023.

\bibitem[Fan et~al.(2025)Fan, Tong, Zhu, Sinha, Liu, Chen, Rabbat, Ballas, LeCun, Bar, et~al.]{webssl}
David Fan, Shengbang Tong, Jiachen Zhu, Koustuv Sinha, Zhuang Liu, Xinlei Chen, Michael Rabbat, Nicolas Ballas, Yann LeCun, Amir Bar, et~al.
\newblock Scaling language-free visual representation learning.
\newblock In \emph{ICCV}, 2025.

\bibitem[Feichtenhofer et~al.(2022)Feichtenhofer, Li, He, et~al.]{maest}
Christoph Feichtenhofer, Yanghao Li, Kaiming He, et~al.
\newblock Masked autoencoders as spatiotemporal learners.
\newblock In \emph{NeurIPS}, 2022.

\bibitem[Fu et~al.(2024)Fu, Lian, Wang, Shi, Wang, Yala, Darrell, Efros, and Goldberg]{crossmae}
Letian Fu, Long Lian, Renhao Wang, Baifeng Shi, Xudong Wang, Adam Yala, Trevor Darrell, Alexei~A Efros, and Ken Goldberg.
\newblock Rethinking patch dependence for masked autoencoders.
\newblock \emph{arXiv:2401.14391}, 2024.

\bibitem[Gao et~al.(2024)Gao, Geng, Zhang, Ma, Fang, Zhang, Li, and Qiao]{gao2024clip}
Peng Gao, Shijie Geng, Renrui Zhang, Teli Ma, Rongyao Fang, Yongfeng Zhang, Hongsheng Li, and Yu Qiao.
\newblock Clip-adapter: Better vision-language models with feature adapters.
\newblock \emph{IJCV}, 2024.

\bibitem[Geiger et~al.(2012)Geiger, Lenz, and Urtasun]{kitti}
Andreas Geiger, Philip Lenz, and Raquel Urtasun.
\newblock Are we ready for autonomous driving? the kitti vision benchmark suite.
\newblock In \emph{CVPR}, 2012.

\bibitem[Hadsell et~al.(2006)Hadsell, Chopra, and LeCun]{Hadsell2006}
Raia Hadsell, Sumit Chopra, and Yann LeCun.
\newblock Dimensionality reduction by learning an invariant mapping.
\newblock In \emph{CVPR}, 2006.

\bibitem[He et~al.(2016)He, Zhang, Ren, and Sun]{resnet}
Kaiming He, Xiangyu Zhang, Shaoqing Ren, and Jian Sun.
\newblock Deep residual learning for image recognition.
\newblock In \emph{CVPR}, 2016.

\bibitem[He et~al.(2019)He, Girshick, and Doll{\'a}r]{he2019rethinking}
Kaiming He, Ross Girshick, and Piotr Doll{\'a}r.
\newblock Rethinking imagenet pre-training.
\newblock In \emph{ICCV}, 2019.

\bibitem[He et~al.(2020)He, Fan, Wu, Xie, and Girshick]{moco}
Kaiming He, Haoqi Fan, Yuxin Wu, Saining Xie, and Ross Girshick.
\newblock Momentum contrast for unsupervised visual representation learning.
\newblock In \emph{CVPR}, 2020.

\bibitem[He et~al.(2022)He, Chen, Xie, Li, Doll{\'a}r, and Girshick]{mae}
Kaiming He, Xinlei Chen, Saining Xie, Yanghao Li, Piotr Doll{\'a}r, and Ross Girshick.
\newblock Masked autoencoders are scalable vision learners.
\newblock In \emph{CVPR}, 2022.

\bibitem[Henaff(2020)]{cpc}
Olivier Henaff.
\newblock Data-efficient image recognition with contrastive predictive coding.
\newblock In \emph{ICML}, 2020.

\bibitem[Hu et~al.(2022)Hu, Debnath, Xie, and Chen]{longmae}
Ronghang Hu, Shoubhik Debnath, Saining Xie, and Xinlei Chen.
\newblock Exploring long-sequence masked autoencoders.
\newblock \emph{arXiv:2210.07224}, 2022.

\bibitem[Huang et~al.(2016)Huang, Sun, Liu, Sedra, and Weinberger]{huang2016deep}
Gao Huang, Yu Sun, Zhuang Liu, Daniel Sedra, and Kilian~Q Weinberger.
\newblock Deep networks with stochastic depth.
\newblock In \emph{ECCV}, 2016.

\bibitem[Hurst et~al.(2024)Hurst, Lerer, Goucher, Perelman, Ramesh, Clark, Ostrow, Welihinda, Hayes, Radford, et~al.]{gpt4o}
Aaron Hurst, Adam Lerer, Adam~P Goucher, Adam Perelman, Aditya Ramesh, Aidan Clark, AJ Ostrow, Akila Welihinda, Alan Hayes, Alec Radford, et~al.
\newblock Gpt-4o system card.
\newblock \emph{arXiv:2410.21276}, 2024.

\bibitem[Kakogeorgiou et~al.(2022)Kakogeorgiou, Gidaris, Psomas, Avrithis, Bursuc, Karantzalos, and Komodakis]{kakogeorgiou2022hide}
Ioannis Kakogeorgiou, Spyros Gidaris, Bill Psomas, Yannis Avrithis, Andrei Bursuc, Konstantinos Karantzalos, and Nikos Komodakis.
\newblock What to hide from your students: Attention-guided masked image modeling.
\newblock In \emph{ECCV}, 2022.

\bibitem[Keetha et~al.(2025)Keetha, M{\"u}ller, Sch{\"o}nberger, Porzi, Zhang, Fischer, Knapitsch, Zauss, Weber, Antunes, et~al.]{mapanything}
Nikhil Keetha, Norman M{\"u}ller, Johannes Sch{\"o}nberger, Lorenzo Porzi, Yuchen Zhang, Tobias Fischer, Arno Knapitsch, Duncan Zauss, Ethan Weber, Nelson Antunes, et~al.
\newblock Mapanything: Universal feed-forward metric 3d reconstruction.
\newblock \emph{arXiv:2509.13414}, 2025.

\bibitem[Kirillov et~al.(2023)Kirillov, Mintun, Ravi, Mao, Rolland, Gustafson, Xiao, Whitehead, Berg, Lo, Dollár, and Girshick]{sam1}
Alexander Kirillov, Eric Mintun, Nikhila Ravi, Hanzi Mao, Chloe Rolland, Laura Gustafson, Tete Xiao, Spencer Whitehead, Alexander~C. Berg, Wan-Yen Lo, Piotr Dollár, and Ross Girshick.
\newblock Segment anything.
\newblock In \emph{CVPR}, 2023.

\bibitem[Krizhevsky et~al.(2012)Krizhevsky, Sutskever, and Hinton]{alexnet}
Alex Krizhevsky, Ilya Sutskever, and Geoffrey~E Hinton.
\newblock Imagenet classification with deep convolutional neural networks.
\newblock In \emph{NeurIPS}, 2012.

\bibitem[Liang et~al.(2023)Liang, Wu, Dai, Li, Zhao, Zhang, Zhang, Vajda, and Marculescu]{liang2023open}
Feng Liang, Bichen Wu, Xiaoliang Dai, Kunpeng Li, Yinan Zhao, Hang Zhang, Peizhao Zhang, Peter Vajda, and Diana Marculescu.
\newblock Open-vocabulary semantic segmentation with mask-adapted clip.
\newblock In \emph{CVPR}, 2023.

\bibitem[Liu and He(2025)]{liu2024decade}
Zhuang Liu and Kaiming He.
\newblock A decade's battle on dataset bias: Are we there yet?
\newblock In \emph{ICLR}, 2025.

\bibitem[Majumdar et~al.(2023)Majumdar, Yadav, Arnaud, Ma, Chen, Silwal, Jain, Berges, Wu, Vakil, et~al.]{cortexbench}
Arjun Majumdar, Karmesh Yadav, Sergio Arnaud, Jason Ma, Claire Chen, Sneha Silwal, Aryan Jain, Vincent-Pierre Berges, Tingfan Wu, Jay Vakil, et~al.
\newblock Where are we in the search for an artificial visual cortex for embodied intelligence?
\newblock In \emph{NeurIPS}, 2023.

\bibitem[Nair et~al.(2022)Nair, Rajeswaran, Kumar, Finn, and Gupta]{nair2022r3m}
Suraj Nair, Aravind Rajeswaran, Vikash Kumar, Chelsea Finn, and Abhinav Gupta.
\newblock R3m: A universal visual representation for robot manipulation.
\newblock In \emph{CoRL}, 2022.

\bibitem[Neuhold et~al.(2017)Neuhold, Ollmann, Rota~Bulo, and Kontschieder]{mapillary}
Gerhard Neuhold, Tobias Ollmann, Samuel Rota~Bulo, and Peter Kontschieder.
\newblock The mapillary vistas dataset for semantic understanding of street scenes.
\newblock In \emph{ICCV}, 2017.

\bibitem[Noroozi and Favaro(2016)]{jigsaw}
Mehdi Noroozi and Paolo Favaro.
\newblock Unsupervised learning of visual representations by solving jigsaw puzzles.
\newblock In \emph{ECCV}, 2016.

\bibitem[Oquab et~al.(2024)Oquab, Darcet, Moutakanni, Vo, Szafraniec, Khalidov, Fernandez, Haziza, Massa, El-Nouby, et~al.]{dinov2}
Maxime Oquab, Timoth{\'e}e Darcet, Th{\'e}o Moutakanni, Huy Vo, Marc Szafraniec, Vasil Khalidov, Pierre Fernandez, Daniel Haziza, Francisco Massa, Alaaeldin El-Nouby, et~al.
\newblock Dinov2: Learning robust visual features without supervision.
\newblock \emph{TMLR}, 2024.

\bibitem[Pathak et~al.(2016)Pathak, Krahenbuhl, Donahue, Darrell, and Efros]{pathak2016context}
Deepak Pathak, Philipp Krahenbuhl, Jeff Donahue, Trevor Darrell, and Alexei~A Efros.
\newblock Context encoders: Feature learning by inpainting.
\newblock In \emph{CVPR}, 2016.

\bibitem[Radford et~al.(2021)Radford, Kim, Hallacy, Ramesh, Goh, Agarwal, Sastry, Askell, Mishkin, Clark, et~al.]{clip}
Alec Radford, Jong~Wook Kim, Chris Hallacy, Aditya Ramesh, Gabriel Goh, Sandhini Agarwal, Girish Sastry, Amanda Askell, Pamela Mishkin, Jack Clark, et~al.
\newblock Learning transferable visual models from natural language supervision.
\newblock In \emph{ICML}, 2021.

\bibitem[Ranftl et~al.(2021)Ranftl, Bochkovskiy, and Koltun]{dpt}
Ren{\'e} Ranftl, Alexey Bochkovskiy, and Vladlen Koltun.
\newblock Vision transformers for dense prediction.
\newblock In \emph{ICCV}, 2021.

\bibitem[Ravi et~al.(2025)Ravi, Gabeur, Hu, Hu, Ryali, Ma, Khedr, R{\"a}dle, Rolland, Gustafson, et~al.]{sam2}
Nikhila Ravi, Valentin Gabeur, Yuan-Ting Hu, Ronghang Hu, Chaitanya Ryali, Tengyu Ma, Haitham Khedr, Roman R{\"a}dle, Chloe Rolland, Laura Gustafson, et~al.
\newblock Sam 2: Segment anything in images and videos.
\newblock In \emph{ICLR}, 2025.

\bibitem[Sablayrolles et~al.(2019)Sablayrolles, Douze, Schmid, and J{\'e}gou]{koleo}
Alexandre Sablayrolles, Matthijs Douze, Cordelia Schmid, and Herv{\'e} J{\'e}gou.
\newblock Spreading vectors for similarity search.
\newblock In \emph{ICLR}, 2019.

\bibitem[Schops et~al.(2017)Schops, Schonberger, Galliani, Sattler, Schindler, Pollefeys, and Geiger]{eth3d}
Thomas Schops, Johannes~L Schonberger, Silvano Galliani, Torsten Sattler, Konrad Schindler, Marc Pollefeys, and Andreas Geiger.
\newblock A multi-view stereo benchmark with high-resolution images and multi-camera videos.
\newblock In \emph{CVPR}, 2017.

\bibitem[Shang et~al.(2024)Shang, Schmeckpeper, May, Minniti, Kelestemur, Watkins, and Herlant]{shang2024theia}
Jinghuan Shang, Karl Schmeckpeper, Brandon~B May, Maria~Vittoria Minniti, Tarik Kelestemur, David Watkins, and Laura Herlant.
\newblock Theia: Distilling diverse vision foundation models for robot learning.
\newblock \emph{arXiv:2407.20179}, 2024.

\bibitem[Shin et~al.(2024)Shin, Lee, Lee, and Lee]{shin2024self}
Jeongwoo Shin, Inseo Lee, Junho Lee, and Joonseok Lee.
\newblock Self-guided masked autoencoder.
\newblock In \emph{NeurIPS}, 2024.

\bibitem[Silberman et~al.(2012)Silberman, Hoiem, Kohli, and Fergus]{nyuv2}
Nathan Silberman, Derek Hoiem, Pushmeet Kohli, and Rob Fergus.
\newblock Indoor segmentation and support inference from rgbd images.
\newblock In \emph{ECCV}, 2012.

\bibitem[Sim{\'e}oni et~al.(2025)Sim{\'e}oni, Vo, Seitzer, Baldassarre, Oquab, Jose, Khalidov, Szafraniec, Yi, Ramamonjisoa, et~al.]{dinov3}
Oriane Sim{\'e}oni, Huy~V Vo, Maximilian Seitzer, Federico Baldassarre, Maxime Oquab, Cijo Jose, Vasil Khalidov, Marc Szafraniec, Seungeun Yi, Micha{\"e}l Ramamonjisoa, et~al.
\newblock Dinov3.
\newblock \emph{arXiv:2508.10104}, 2025.

\bibitem[Simonyan and Zisserman(2015)]{vggnet}
Karen Simonyan and Andrew Zisserman.
\newblock Very deep convolutional networks for large-scale image recognition.
\newblock In \emph{ICLR}, 2015.

\bibitem[Thomee et~al.(2016)Thomee, Shamma, Friedland, Elizalde, Ni, Poland, Borth, and Li]{yfcc}
Bart Thomee, David~A Shamma, Gerald Friedland, Benjamin Elizalde, Karl Ni, Douglas Poland, Damian Borth, and Li-Jia Li.
\newblock Yfcc100m: The new data in multimedia research.
\newblock \emph{Communications of the ACM}, 2016.

\bibitem[Tian et~al.(2020)Tian, Krishnan, and Isola]{cmc}
Yonglong Tian, Dilip Krishnan, and Phillip Isola.
\newblock Contrastive multiview coding.
\newblock In \emph{ECCV}, 2020.

\bibitem[Tong et~al.(2022)Tong, Song, Wang, and Wang]{videomae}
Zhan Tong, Yibing Song, Jue Wang, and Limin Wang.
\newblock Videomae: Masked autoencoders are data-efficient learners for self-supervised video pre-training.
\newblock In \emph{NeurIPS}, 2022.

\bibitem[Tschannen et~al.(2025)Tschannen, Gritsenko, Wang, Naeem, Alabdulmohsin, Parthasarathy, Evans, Beyer, Xia, Mustafa, et~al.]{siglip2}
Michael Tschannen, Alexey Gritsenko, Xiao Wang, Muhammad~Ferjad Naeem, Ibrahim Alabdulmohsin, Nikhil Parthasarathy, Talfan Evans, Lucas Beyer, Ye Xia, Basil Mustafa, et~al.
\newblock Siglip 2: Multilingual vision-language encoders with improved semantic understanding, localization, and dense features.
\newblock \emph{arXiv:2502.14786}, 2025.

\bibitem[Vasiljevic et~al.(2019)Vasiljevic, Kolkin, Zhang, Luo, Wang, Dai, Daniele, Mostajabi, Basart, Walter, et~al.]{diode}
Igor Vasiljevic, Nick Kolkin, Shanyi Zhang, Ruotian Luo, Haochen Wang, Falcon~Z Dai, Andrea~F Daniele, Mohammadreza Mostajabi, Steven Basart, Matthew~R Walter, et~al.
\newblock Diode: A dense indoor and outdoor depth dataset.
\newblock \emph{arXiv:1908.00463}, 2019.

\bibitem[Villalobos et~al.(2024)Villalobos, Ho, Sevilla, Besiroglu, Heim, and Hobbhahn]{villalobos2024position}
Pablo Villalobos, Anson Ho, Jaime Sevilla, Tamay Besiroglu, Lennart Heim, and Marius Hobbhahn.
\newblock Position: Will we run out of data? limits of llm scaling based on human-generated data.
\newblock In \emph{ICML}, 2024.

\bibitem[Vincent et~al.(2008)Vincent, Larochelle, Bengio, and Manzagol]{vincent2008extracting}
Pascal Vincent, Hugo Larochelle, Yoshua Bengio, and Pierre-Antoine Manzagol.
\newblock Extracting and composing robust features with denoising autoencoders.
\newblock In \emph{ICML}, 2008.

\bibitem[Vo et~al.(2024)Vo, Khalidov, Darcet, Moutakanni, Smetanin, Szafraniec, Touvron, Couprie, Oquab, Joulin, et~al.]{vo2024automatic}
Huy~V Vo, Vasil Khalidov, Timoth{\'e}e Darcet, Th{\'e}o Moutakanni, Nikita Smetanin, Marc Szafraniec, Hugo Touvron, Camille Couprie, Maxime Oquab, Armand Joulin, et~al.
\newblock Automatic data curation for self-supervised learning: A clustering-based approach.
\newblock \emph{TMLR}, 2024.

\bibitem[Wang et~al.(2021)Wang, Zheng, Ma, Lu, and Zhong]{loveda}
Junjue Wang, Zhuo Zheng, Ailong Ma, Xiaoyan Lu, and Yanfei Zhong.
\newblock Loveda: A remote sensing land-cover dataset for domain adaptive semantic segmentation.
\newblock In \emph{NeurIPS}, 2021.

\bibitem[Wang et~al.(2024)Wang, Leroy, Cabon, Chidlovskii, and Revaud]{dust3r}
Shuzhe Wang, Vincent Leroy, Yohann Cabon, Boris Chidlovskii, and Jerome Revaud.
\newblock Dust3r: Geometric 3d vision made easy.
\newblock In \emph{CVPR}, 2024.

\bibitem[Wang et~al.(2020)Wang, Zhu, Wang, Hu, Qiu, Wang, Hu, Kapoor, and Scherer]{tartanair}
Wenshan Wang, Delong Zhu, Xiangwei Wang, Yaoyu Hu, Yuheng Qiu, Chen Wang, Yafei Hu, Ashish Kapoor, and Sebastian Scherer.
\newblock Tartanair: A dataset to push the limits of visual slam.
\newblock In \emph{IROS}, 2020.

\bibitem[Wang et~al.(2025)Wang, Zhou, He, Huang, Yang, Zhang, Cheng, Ji, Jin, Zhao, et~al.]{wang2025spatialclip}
Zehan Wang, Sashuai Zhou, Shaoxuan He, Haifeng Huang, Lihe Yang, Ziang Zhang, Xize Cheng, Shengpeng Ji, Tao Jin, Hengshuang Zhao, et~al.
\newblock Spatialclip: Learning 3d-aware image representations from spatially discriminative language.
\newblock In \emph{CVPR}, 2025.

\bibitem[Wiedemer et~al.(2025)Wiedemer, Li, Vicol, Gu, Matarese, Swersky, Kim, Jaini, and Geirhos]{wiedemer2025video}
Thadd{\"a}us Wiedemer, Yuxuan Li, Paul Vicol, Shixiang~Shane Gu, Nick Matarese, Kevin Swersky, Been Kim, Priyank Jaini, and Robert Geirhos.
\newblock Video models are zero-shot learners and reasoners.
\newblock \emph{arXiv:2509.20328}, 2025.

\bibitem[Wu et~al.(2018)Wu, Xiong, Yu, and Lin]{instdisc}
Zhirong Wu, Yuanjun Xiong, Stella~X Yu, and Dahua Lin.
\newblock Unsupervised feature learning via non-parametric instance discrimination.
\newblock In \emph{CVPR}, 2018.

\bibitem[Xie et~al.(2022)Xie, Zhang, Cao, Lin, Bao, Yao, Dai, and Hu]{simmim}
Zhenda Xie, Zheng Zhang, Yue Cao, Yutong Lin, Jianmin Bao, Zhuliang Yao, Qi Dai, and Han Hu.
\newblock Simmim: A simple framework for masked image modeling.
\newblock In \emph{CVPR}, 2022.

\bibitem[Xu et~al.(2024)Xu, Xie, Tan, Huang, Howes, Sharma, Li, Ghosh, Zettlemoyer, and Feichtenhofer]{metaclip}
Hu Xu, Saining Xie, Xiaoqing~Ellen Tan, Po-Yao Huang, Russell Howes, Vasu Sharma, Shang-Wen Li, Gargi Ghosh, Luke Zettlemoyer, and Christoph Feichtenhofer.
\newblock Demystifying clip data.
\newblock In \emph{ICLR}, 2024.

\bibitem[Yang et~al.(2024)Yang, Kang, Huang, Zhao, Xu, Feng, and Zhao]{dav2}
Lihe Yang, Bingyi Kang, Zilong Huang, Zhen Zhao, Xiaogang Xu, Jiashi Feng, and Hengshuang Zhao.
\newblock Depth anything v2.
\newblock In \emph{NeurIPS}, 2024.

\bibitem[Yeshwanth et~al.(2023)Yeshwanth, Liu, Nie{\ss}ner, and Dai]{scannet++}
Chandan Yeshwanth, Yueh-Cheng Liu, Matthias Nie{\ss}ner, and Angela Dai.
\newblock Scannet++: A high-fidelity dataset of 3d indoor scenes.
\newblock In \emph{ICCV}, 2023.

\bibitem[Zhai et~al.(2023)Zhai, Mustafa, Kolesnikov, and Beyer]{siglip}
Xiaohua Zhai, Basil Mustafa, Alexander Kolesnikov, and Lucas Beyer.
\newblock Sigmoid loss for language image pre-training.
\newblock In \emph{ICCV}, 2023.

\bibitem[Zhang et~al.(2024)Zhang, Zhang, Dong, Zang, and Wang]{longclip}
Beichen Zhang, Pan Zhang, Xiaoyi Dong, Yuhang Zang, and Jiaqi Wang.
\newblock Long-clip: Unlocking the long-text capability of clip.
\newblock In \emph{ECCV}, 2024.

\bibitem[Zhang et~al.(2025{\natexlab{a}})Zhang, Moing, Koppula, Rocco, Momeni, Xie, Sun, Sukthankar, Barral, Hadsell, Ghahramani, Zisserman, Zhang, and Sajjadi]{d4rt}
Chuhan Zhang, Guillaume~Le Moing, Skanda Koppula, Ignacio Rocco, Liliane Momeni, Junyu Xie, Shuyang Sun, Rahul Sukthankar, Joëlle~K. Barral, Raia Hadsell, Zoubin Ghahramani, Andrew Zisserman, Junlin Zhang, and Mehdi S.~M. Sajjadi.
\newblock Efficiently reconstructing dynamic scenes one d4rt at a time.
\newblock \emph{arXiv:2512.08924}, 2025{\natexlab{a}}.

\bibitem[Zhang et~al.(2022)Zhang, Wang, and Wang]{zhang2022mask}
Qi Zhang, Yifei Wang, and Yisen Wang.
\newblock How mask matters: Towards theoretical understandings of masked autoencoders.
\newblock In \emph{NeurIPS}, 2022.

\bibitem[Zhang et~al.(2025{\natexlab{b}})Zhang, Keetha, Lyu, Jhamb, Chen, Qiu, Karhade, Jha, Hu, Ramanan, et~al.]{zhang2025ufm}
Yuchen Zhang, Nikhil Keetha, Chenwei Lyu, Bhuvan Jhamb, Yutian Chen, Yuheng Qiu, Jay Karhade, Shreyas Jha, Yaoyu Hu, Deva Ramanan, et~al.
\newblock Ufm: A simple path towards unified dense correspondence with flow.
\newblock \emph{arXiv:2506.09278}, 2025{\natexlab{b}}.

\bibitem[Zhou et~al.(2017)Zhou, Zhao, Puig, Fidler, Barriuso, and Torralba]{ade20k}
Bolei Zhou, Hang Zhao, Xavier Puig, Sanja Fidler, Adela Barriuso, and Antonio Torralba.
\newblock Scene parsing through ade20k dataset.
\newblock In \emph{CVPR}, 2017.

\bibitem[Zhou et~al.(2022)Zhou, Wei, Wang, Shen, Xie, Yuille, and Kong]{ibot}
Jinghao Zhou, Chen Wei, Huiyu Wang, Wei Shen, Cihang Xie, Alan Yuille, and Tao Kong.
\newblock ibot: Image bert pre-training with online tokenizer.
\newblock In \emph{ICLR}, 2022.

\end{thebibliography}
